\documentclass{article}

\PassOptionsToPackage{numbers, compress}{natbib}

\usepackage[main, final]{neurips_2026}

\usepackage{amsmath}
\usepackage{algorithm}
\usepackage{algorithmic}
\usepackage[table,dvipsnames]{xcolor}
\usepackage{multirow}
\usepackage{setspace}
\usepackage{makecell}
\usepackage{placeins}
\usepackage{gradient-text}

\usepackage[utf8]{inputenc} %
\usepackage[T1]{fontenc}    %
\usepackage{hyperref}       %
\usepackage{url}            %
\usepackage{booktabs}       %
\usepackage{amsfonts}       %
\usepackage{nicefrac}       %
\usepackage{microtype}      %

\def\rvd{{\mathbf{d}}}

\def\rvm{{\mathbf{m}}}

\def\rvp{{\mathbf{p}}}

\def\rvv{{\mathbf{v}}}

\def\rvx{{\mathbf{x}}}

\DeclareMathAlphabet{\mathsfit}{\encodingdefault}{\sfdefault}{m}{sl}
\SetMathAlphabet{\mathsfit}{bold}{\encodingdefault}{\sfdefault}{bx}{n}

\usepackage[normalem]{ulem}
\usepackage{inconsolata}
\usepackage{xspace}

\newcommand{\modelname}{\textsc{CovRAG}\xspace}  %

\usepackage{graphicx}
\usepackage{enumitem}
\usepackage{caption}
\usepackage{wrapfig}
\usepackage{float}

\title{Retrieve What’s Missing: Coverage-Maximizing Retrieval for Consistent Long Video Generation}

\newcommand{\ProjectURL}[1]{%
  {\hypersetup{hidelinks}%
   \href{#1}{{%
     \fontencoding{T1}\fontfamily{cmtt}%
     \fontseries{m}\fontshape{n}\selectfont
     \gradientRGB{#1}{143,21,56}{232,62,121}%
   }}}%
}
\author{%
  Minseok Joo$^{1}$ \quad
  Dogyun Park$^{1}$ \quad
  Taehoon Lee$^{2}$ \quad
  Kyujin Lee$^{2}$ \quad
  Hyunwoo J. Kim$^{2}$\thanks{Corresponding author.}\\[0.5em]
  {\normalfont
    $^{1}$Korea University \qquad $^{2}$KAIST
  }
  \\[0.65em]
  \ProjectURL{https://mlvlab.github.io/CovRAG}%
}

\begin{document}

\begin{center}
    \maketitle 
    \vspace{-8mm}
    \captionsetup{type=figure}
    \includegraphics[width=0.98\textwidth]{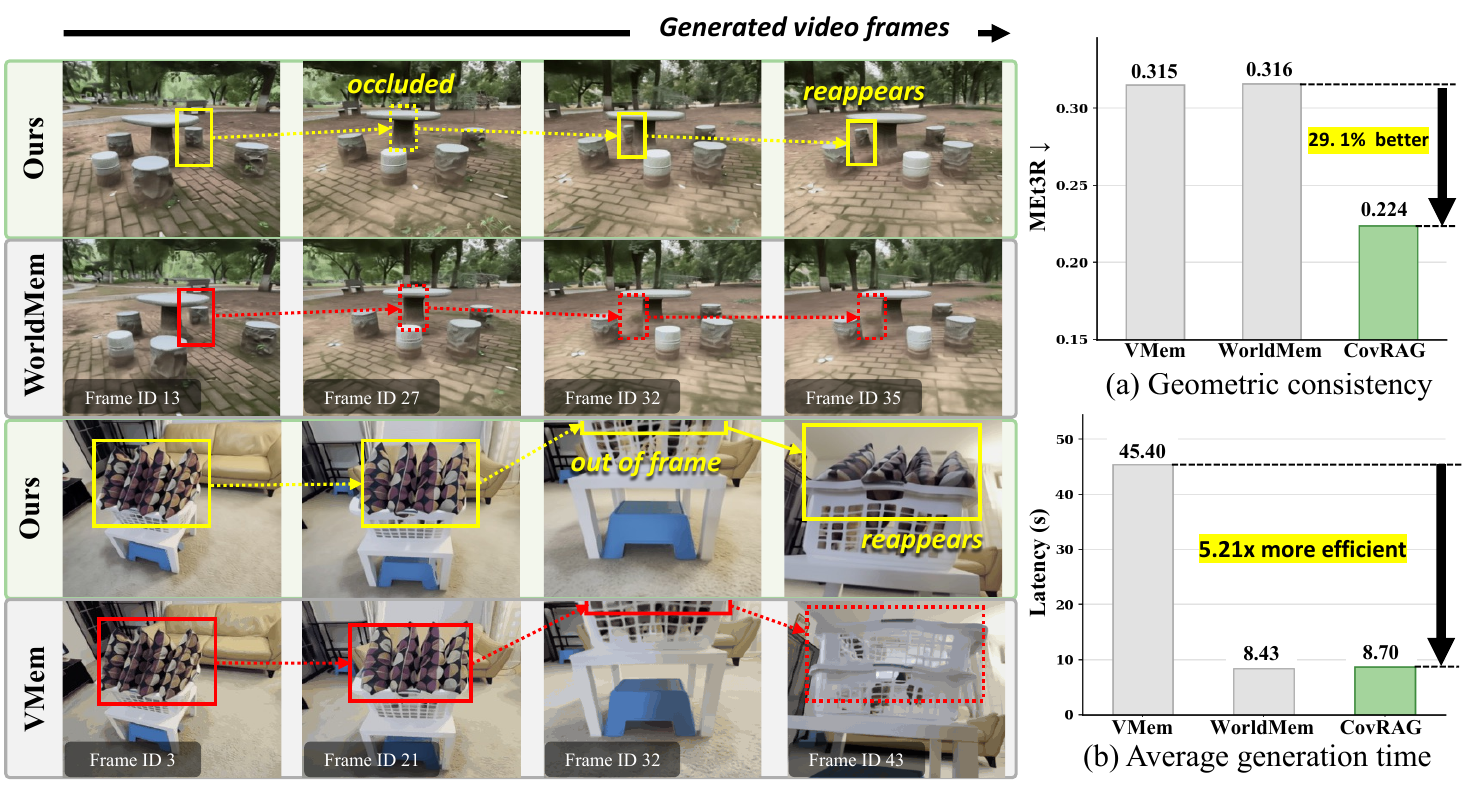}
    \caption{\textbf{Consistent scenes with \modelname(Ours).} By maximizing residual coverage, \modelname retrieves complementary frames to preserve reappearing content, achieving a better consistency--efficiency trade-off than FoV-based retrieval (WorldMem) and global 3D scene memory (VMem).}
    \label{fig:figure1}
\end{center}

\begin{abstract}
\vspace{-2mm}
Long-horizon video generation is a key capability of \emph{video world models}, yet maintaining geometric consistency remains challenging.
Memory-augmented generative models address this by retrieving historical frames, but their effectiveness depends on two key design choices: what 3D-geometric evidence should represent past observations, and how memory frames should be selected from this evidence. 
Pose-based field-of-view overlap provides lightweight but coarse geometric evidence, while explicit 3D scene memories offer finer spatial information at the cost of geometry estimation, alignment, and incremental updates over long rollouts.
We propose \emph{\textbf{Cov}erage-Maximizing \textbf{R}etrieval-\textbf{A}ugmented \textbf{G}eneration} (\modelname), a depth-based memory retrieval framework that uses pretrained 3D priors to construct a \emph{target-view coverage map} as lightweight 3D memory evidence. 
For frame selection, \modelname maximizes \emph{residual coverage gain}, iteratively retrieving frames that explain target-view regions not covered by the temporal context or previously selected memories.
To improve scalability in long-video generation, we introduce sliding-window depth caching for efficient geometry estimation. 
Experiments on RealEstate10K and DL3DV10K show that \modelname improves long-horizon geometric consistency over the evaluated memory-retrieval baselines while maintaining low retrieval latency.
\end{abstract}
\section{Introduction}
Simulating the visual world is a central goal of computer vision.
Recent \emph{video world models} pursue this goal by generating videos conditioned on camera motion or other control signals~\cite{sun2026worldplay,nvidia2025cosmosworldfoundationmodel,seo2026grounding,he2025matrix}.
For such models, maintaining a consistent 3D scene across views over long horizons is essential for robotic simulation~\cite{wenvidman,wolf2025diffusion,YangDGTKSA24,zhu2025irasim}, 3D content creation~\cite{yu2024wonderjourney,he2025cameractrl,bai2025recammaster,gao2024cat3d,yu2025viewcrafter,yu2025wonderworld}, and interactive game engines~\cite{feng2024matrix,yu2025gamefactory,valevski2024diffusion,bruce2024genie}.
Despite this progress, maintaining \emph{long-term geometric consistency} remains challenging.
Most autoregressive video diffusion models condition on a limited window of recent frames and thus struggle to preserve structural and semantic details from earlier observations outside that window.
When the camera revisits a previously observed region, the model may fail to reproduce geometry and appearance consistent with the earlier observation, leading to spatial drift, flickering, and inconsistent scene structure.
A promising direction is \emph{memory-augmented video generation}~\cite{xiao2025worldmem,yu2025cam,li2025vmem,chen2025learningworldmodelsinteractive,henschel2025streamingt2v,gao2026memcam}, which stores historical frames or states and retrieves a subset of them to condition future generation.
This approach provides access to a long history through a small set of conditioning frames, allowing retrieval to be improved without increasing the generator's context length.
However, the effectiveness of memory-augmented generation depends critically on the retrieval mechanism.
In particular, retrieval must answer two key questions:
\emph{(i) What form of 3D-aware evidence best captures the geometric relevance of past observations?}
and
\emph{(ii) How should frames be selected based on this evidence?}

Existing approaches use different forms of geometric evidence, with trade-offs between spatial precision and memory-maintenance cost.
Some approaches, such as WorldMem~\cite{xiao2025worldmem}, use camera poses or field-of-view (FoV) overlap as lightweight 3D-aware evidence.
This is efficient and easy to scale, but it provides only a coarse view-level proxy for geometric relevance.
Pose-based FoV overlap does not explicitly account for scene occlusions or identify which target-view pixels are geometrically supported by a memory frame.
This limits its ability to identify historical evidence for regions that are occluded or outside the field of view in the current temporal context, as illustrated by the reappearance examples in Figure~\ref{fig:figure1}.
Methods such as VMem~\cite{li2025vmem} and Spatia~\cite{zhao2026spatia} maintain explicit 3D scene memories to identify relevant historical views.
These scene memories provide fine-grained geometric information, but maintaining them requires geometry estimation, alignment, and incremental updates, which add computational overhead and can limit scalability over long rollouts.
For the second question, scoring candidates independently and retrieving the top-$n$ can identify individually relevant frames without ensuring that the retrieved set is \emph{collectively informative}.
Multiple frames may redundantly explain the same region, while other regions remain unsupported.
This redundancy is especially harmful under tight context-length constraints and compute budgets, as redundant references occupy conditioning slots that could otherwise be used for complementary evidence.

\begin{figure*}[!t]
    \centering
    \includegraphics[width=0.95\textwidth]{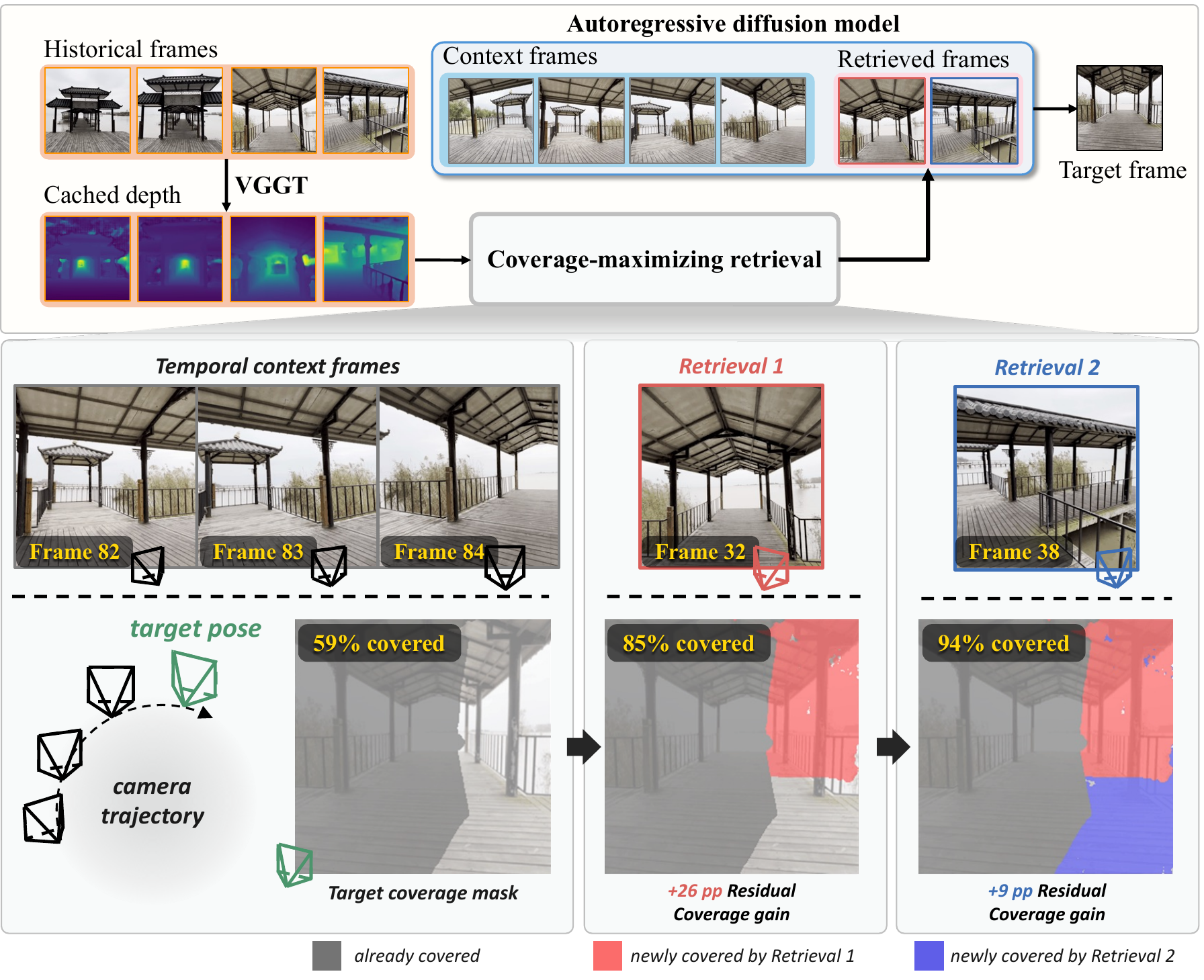}
    \caption{\textbf{Overview of \modelname.} During autoregressive video generation, \modelname augments the diffusion model with retrieved historical frames that provide complementary geometric evidence for the target view. It identifies regions already supported by the temporal context and retrieves additional memories (denoted by red and blue regions) that cover the remaining target-view regions, enabling more consistent long-horizon generation. \textit{pp} denotes percentage points.}
    \label{fig:figure2}
\vspace{-5mm}
\end{figure*}

Thus, we propose \textbf{Coverage-Maximizing Retrieval} (\modelname), a depth-based retrieval framework, illustrated in Figure~\ref{fig:figure2}, to address these challenges.
It requires no additional training when the generator already supports historical frame conditioning.
For memory evidence, \modelname uses depth predicted by a pretrained 3D prior model~\cite{wang2025vggt} and camera poses to construct a \emph{target-view coverage map}.
Concretely, each historical frame is warped to the target viewpoint using its predicted depth and camera pose, and the resulting target-view coverage map estimates which target-view pixels are geometrically supported by that frame.
This depth-based evidence provides a practical middle ground by supporting pixel-level coverage reasoning without maintaining a global 3D scene representation.
For frame selection, \modelname retrieves frames to maximize \emph{residual coverage gain}, the number of additional target-view pixels a candidate covers beyond the temporal context and already-selected references.
This prioritizes complementary geometric evidence within a fixed retrieval budget.
To keep geometry estimation scalable over long rollouts, we introduce a \emph{sliding-window depth-caching strategy} that confines geometry inference to a fixed window and reuses cached depth maps thereafter, keeping per-step geometry inference cost nearly constant as the generated sequence grows.

We evaluate \modelname and memory-retrieval baselines under a common DFoT~\cite{song2025historyguidedvideodiffusion} backbone on RealEstate10K~\cite{zhou2018stereo} and DL3DV10K~\cite{ling2024dl3dv}, which together cover diverse 3D scenes with varied camera trajectories and substantial spatial extent.
Compared with FoV-based retrieval (WorldMem), \modelname reduces the geometric consistency error measured by MEt3R by up to \textbf{29\%}, while achieving an approximately \textbf{5$\times$} generation speedup over an explicit 3D memory-based baseline (VMem), measured by average time per generated frame (Figure~\ref{fig:figure1}).
These results indicate that \modelname is an effective and efficient retrieval method for 3D-consistent video generation.

Our \textbf{contributions} are summarized as follows:
\begin{itemize}[leftmargin=1em,itemsep=0.2em, topsep=0.2em, parsep=0pt]
\item We propose \emph{Coverage-Maximizing Retrieval} (\modelname), a depth-based memory-retrieval framework to improve geometric consistency in long-horizon autoregressive video generation.

\item We introduce \emph{target-view coverage maps}, a lightweight 3D-aware memory evidence representation that uses pretrained 3D priors to estimate pixel-level geometric support from historical frames without maintaining a global 3D scene representation.

\item We formulate memory selection as maximizing \emph{residual coverage gain} beyond both the temporal context and already selected references and use greedy selection to retrieve complementary geometric evidence under a limited conditioning budget.

\item We enable scalable depth-based coverage retrieval with \emph{sliding-window depth caching}, which estimates depth only within a fixed local rollout window and reuses cached depth maps for later coverage computation, keeping per-step geometry inference cost nearly constant as the video grows.

\end{itemize}

\section{Related Work}

\noindent\textbf{Long-horizon autoregressive video generation.}
Diffusion-based video generation has rapidly improved in visual fidelity and temporal coherence~\cite{ho2020denoising,song2020score,kong2024hunyuanvideo,yang2024cogvideox,wan2025wan,park2024ddmi}. 
Long-horizon synthesis is important for interactive exploration, where successive views must remain coherent as users navigate and revisit an environment.
Diffusion Forcing~\cite{chen2025diffusion} enables sequence generation with independent noise levels across frames, while DFoT~\cite{song2025historyguidedvideodiffusion} supports flexible history conditioning and history guidance.
For efficient streaming, CausVid~\cite{yin2025slow} distills bidirectional video diffusion models into few-step causal generators, and AAPT~\cite{lin2025aapt} uses adversarial post-training for one-step autoregressive generation.
Self Forcing~\cite{huang2025self} addresses exposure bias by training on the model's own autoregressive outputs.
FramePack~\cite{zhang2025framepack} compresses historical frame representations according to their importance, allowing more history to fit within a fixed context budget.
Together, these advances improve generation efficiency, history utilization, and rollout stability.
However, extending the generation horizon does not by itself ensure that detailed evidence from earlier views remains available during revisitation.
Persistent memory complements these developments by retaining historical observations beyond the active context.
Our work focuses on selecting complementary historical frames under a limited conditioning budget to support long-horizon geometric consistency.

\textbf{Memory-augmented video generation.}
A growing body of work uses historical observations or geometric representations to guide video generation~\cite{xiao2025worldmem,yu2025cam,li2025vmem,gao2026memcam,cai2025mixture,liu2024reconx,zhang2025world,ren2025gen3c,wu2025spmem,zhao2026spatia}.
These methods differ in their memory representations and conditioning mechanisms, as well as the criteria used to select historical evidence.
GEN3C~\cite{ren2025gen3c} uses renderings of depth-derived 3D caches, while SPMem~\cite{wu2025spmem} combines static point-cloud renderings with recent context and sparse historical references.
MosaicMem~\cite{mosaicmem2026} represents memory as 3D-localized patches and aligns retrieved patches with the target view through positional or latent warping.
For reference selection, WorldMem~\cite{xiao2025worldmem} and Context-as-Memory~\cite{yu2025cam} use pose-based field-of-view (FoV) overlap as an efficient but coarse view-level criterion.
VMem~\cite{li2025vmem} maintains an explicit surfel-indexed scene memory to identify relevant historical views, optimizing newly predicted depth maps for geometric alignment during memory updates.
Spatia~\cite{zhao2026spatia} maintains an updatable global scene point cloud and retrieves references using 3D IoU between registered, view-specific point clouds.
Maintaining these scene memories requires geometry estimation, alignment, and incremental updates, which add computational overhead and can limit scalability over long rollouts.
The retrieved frames should also provide complementary evidence, as independent per-frame ranking can yield redundant references under a limited conditioning budget.
Recent closely related work, including AnchorWeave~\cite{wang2026anchorweave} and Lyra~2.0~\cite{shen2026lyra}, likewise uses coverage-aware retrieval to select complementary memories.
\modelname focuses on inference-time frame selection using cached per-frame depth maps, maximizing additional target-view coverage beyond the temporal context and already-selected references while limiting geometry estimation to a sliding window.

\vspace{-1.5mm}
\section{Preliminaries}
\vspace{-1.5mm}
\noindent\textbf{Notation.}
We use bold lowercase symbols (e.g., $\rvx$) for video tensors, calligraphic symbols (e.g., $\mathcal{C}, \mathcal{M}, \mathcal{H}$) for index sets, subscripts to indicate frame indices, and superscripts for diffusion noise level. Given a frame index set $\mathcal{S}$, we write $\rvx_\mathcal{S}=\{\rvx_j\}_{j \in \mathcal{S}}$ for the subset of frames indexed by $\mathcal{S}$. 

\noindent\textbf{Autoregressive video diffusion.}
Let $\rvx^0 = \{\rvx_f^0\}_{f=1}^F$ denote a clean video with $F$ frames.
Diffusion models generate $\rvx^0$ by starting from Gaussian noise and iteratively denoising through a learned reverse process parameterized by a network~\cite{ho2020denoising,song2020score,lipman2022flow,park2025sprint,park2025blockwise}.
Using a continuous noise level $k\in[0,1]$, where $k=1$ corresponds to pure noise and $k=0$ to the clean video, we write the noised sample as
$\rvx^k = \alpha(k)\,\rvx^0 + \sigma(k)\,\epsilon$ with $\epsilon\sim\mathcal{N}(0,\mathrm{I})$, where $\alpha(k)$ and $\sigma(k)$ are noise schedule functions satisfying $\alpha(0)=1$, $\sigma(0)=0$ and $\alpha(1)=0$, $\sigma(1)=1$.
A common training objective is to learn a noise predictor $\epsilon_\theta$ by minimizing
\begingroup
\setlength{\abovedisplayskip}{1pt}
\setlength{\belowdisplayskip}{1pt}
\begin{equation}
\mathcal{L}_\theta
= \mathbb{E}_{\rvx^0,\epsilon,k}\Big[\big\|\epsilon_\theta(\rvx^k,k)-\epsilon\big\|^2\Big].
\end{equation}
\endgroup
Autoregressive video diffusion extends this framework to long-horizon generation by denoising frame $\rvx_t^k$ conditioned on a temporal context window of previously generated clean frames $\rvx_{\mathcal{C}_t}^0$, where $\mathcal{C}_t = \{t-1, \ldots, t-\delta\}$ is the index set of the most recent $\delta$ frames.
Diffusion Forcing (DF)~\cite{song2025historyguidedvideodiffusion,chen2025diffusion} generalizes this formulation by assigning an independent noise level to each frame, allowing the model to denoise the current frame while conditioning on context frames at arbitrary noise levels.

\textbf{Memory bank with retrieval.} Let $\mathcal{M}_t$ denote the index set of the memory bank at generation frame $t$, and similarly $\mathcal{H}_t$ for retrieved indices.
Memory-augmented video generators~\cite{xiao2025worldmem,yu2025cam,chen2025learningworldmodelsinteractive} maintain an external memory bank $\rvx_{\mathcal{M}_t}$ at current frame $t$ that stores previously generated frames and associated metadata such as camera poses.
Before generating the next frame $\rvx_t^0$, the model retrieves a subset of memory frames $\rvx^0_{\mathcal{H}_t} \subseteq \rvx_{\mathcal{M}_t}^0$,
and conditions the network on $\rvx^0_{\mathcal{H}_t}$, \textit{i.e.}, $\epsilon_\theta(\rvx^k_{t},\,\rvx^0_{\mathcal{C}_t},\,\rvx^0_{\mathcal{H}_t},\,k)$.

\vspace{-1.5mm}
\section{Method}
\vspace{-1.5mm}
Our goal is to retrieve a compact set of historical frames
$\rvx^0_{\mathcal{H}_t} \subseteq \rvx^0_{\mathcal{M}_t}$
that provides useful geometric evidence for generating the target frame $\rvx_t^0$.
The key idea is to identify which regions of the target view are already geometrically supported by the current context, and then retrieve frames that support the remaining uncovered regions.
Our method consists of three main components.
First, in Sec.~\ref{sec:target_view_coverage}, we construct \emph{target-view coverage maps} by warping context and memory frames into the target viewpoint using estimated depth and camera poses.
A target-view coverage map indicates which target pixels are geometrically supported by a given source frame.
Second, in Sec.~\ref{sec:residual_gain}, we select memory frames according to \emph{residual coverage gain}, which measures how much additional target-view coverage a candidate frame provides beyond the current context and the already-selected memory frames.
This encourages retrieved frames to be complementary rather than redundant.
Finally, in Sec.~\ref{sec:geometry_estimation}, we introduce \emph{sliding-window depth caching} for scalable long-horizon geometry estimation.

\subsection{Target-view coverage maps as 3D memory evidence}
\label{sec:target_view_coverage}

\modelname uses depth maps to estimate which target-view pixels are covered by each historical frame.
For each source frame $i$ and target frame $t$, we construct a \emph{target-view coverage map}
$\rvm_{i\rightarrow t}\in\{0,1\}^{H\times W}$,
where $H$ and $W$ denote the frame height and width.
The map $\rvm_{i\rightarrow t}$ indicates which pixels in the target frame $t$ are geometrically supported by the source frame $i$.
Let $\rvd_i$ be the depth map of source frame $i$, and let
$\rvp_i=[\mathbf{R}_i\mid \mathbf{t}_i]$ and
$\rvp_t=[\mathbf{R}_t\mid \mathbf{t}_t]$
denote the camera poses of the source and target views, respectively.
We use ground-truth camera poses, with translations rescaled using VGGT pose estimates (Appendix~\ref{sec:vggt_scaled}).
For each pixel $\mathbf{u}$ in frame $i$, we back-project it to 3D using its depth $\rvd_i(\mathbf{u})$, transform the resulting 3D point into the target camera coordinate system, and project it onto the target image plane.
If the projected point lands on a valid target pixel $\mathbf{v}$, we mark that pixel as covered:
\begin{equation}
\rvm_{i\rightarrow t}(\mathbf{v}) = 1. 
\label{eq:coverage_map}
\end{equation}
Specifically, for each point with positive target-camera depth, we round its projected coordinates to the nearest target pixel $\rvv$ and mark it as covered if $\rvv \in [1,H]\times[1,W]$.
Otherwise, the target pixel remains uncovered, i.e.,
$\rvm_{i\rightarrow t}(\mathbf{v}) = 0$.
In practice, multiple source pixels may project to the same target pixel; we mark a target pixel as covered if at least one source pixel lands on it.
For the current context window $\mathcal{C}_t$, we aggregate the coverage maps of all context frames into a single \emph{context coverage map}:
\begin{equation}
\rvm^{\mathrm{ctx}}_t
=
\bigvee_{i\in\mathcal{C}_t}
\rvm_{i\rightarrow t},
\label{eq:context_coverage_map}
\end{equation}
where $\vee$ denotes the element-wise logical OR operator.
This map indicates which target-view pixels are already covered by the recent temporal context.
The remaining uncovered pixels define the regions where additional historical memory evidence may be useful.
The target-view coverage map provides a lightweight but spatially fine-grained form of memory evidence.
Unlike FoV overlap, it distinguishes which pixels of the target view are supported by a source frame.
Moreover, it computes target-view coverage directly from cached per-frame depth maps without maintaining a global 3D scene representation.
Thus, depth-based coverage maps provide a practical balance between spatial detail and retrieval efficiency.

\vspace{-1mm}
\subsection{Residual coverage gain for non-redundant retrieval}
\vspace{-1mm}
\label{sec:residual_gain}
Given target-view coverage maps, the next question is how to select memory frames.
A simple strategy is to score each candidate frame independently by its total target-view coverage and retrieve the top-$n$ frames.
However, this can select redundant frames that cover the same target-view regions, leaving other target-view regions uncovered.
Since the number of retrieved frames is limited by the conditioning budget and computational cost, retrieval should favor frames that provide complementary evidence.
To this end, \modelname selects frames according to \emph{residual coverage gain}.

\begin{wrapfigure}{R}{0.48\textwidth}
\begin{minipage}{0.48\textwidth}
\input{sec/3_method/algorithm1}
\end{minipage}
\end{wrapfigure}

\noindent\textbf{Residual coverage gain.}
For each candidate memory frame $j\in\mathcal{M}_t$, we compute its residual coverage gain as
\begingroup
\setlength{\abovedisplayskip}{1pt}
\setlength{\belowdisplayskip}{1pt}
\begin{equation}
g_j = \sum_{\mathbf{v}} \left(1-\rvm^{\mathrm{ctx}}_t(\mathbf{v})\right)
\rvm_{j\rightarrow t}(\mathbf{v}).
\label{eq:gain}
\end{equation}
\endgroup
Here, $\rvm^{\mathrm{ctx}}_t$ denotes the accumulated coverage of the temporal context and already-selected memory frames, and $\rvm_{j\rightarrow t}$ denotes the target-view coverage map of candidate frame $j$.
Pixels already covered by the context or selected memories, i.e., $\rvm^{\mathrm{ctx}}_t(\mathbf{v})=1$, do not contribute to the score.
Thus, $g_j$ counts the number of currently uncovered target-view pixels that would become covered if frame $j$ were retrieved.
A higher $g_j$ indicates that frame $j$ covers a larger area of previously uncovered target-view regions; a $g_j$ of zero indicates that the frame provides no additional geometric evidence in terms of target-view coverage.

\noindent\textbf{Retrieval by maximizing residual coverage gain.}
We greedily select $n$ memory frames by maximizing residual coverage gain at each iteration.
We initialize the retrieved set as $\mathcal{H}_t=\emptyset$ and compute the initial context coverage map $\rvm^{\mathrm{ctx}}_t$ from the current context window.
At each iteration, we choose the candidate with the largest residual coverage gain:
\begin{equation}
j^\star = \arg\max_{j\in \mathcal{M}_t\setminus \mathcal{H}_t}
g_j,
\label{eq:argmax}
\end{equation}
and add it to the retrieved set, $\mathcal{H}_t \leftarrow \mathcal{H}_t \cup \{j^\star\}$.
We then update the accumulated coverage map by incorporating the newly selected frame:
\begin{equation}
\rvm^{\mathrm{ctx}}_t
\leftarrow
\rvm^{\mathrm{ctx}}_t
\vee
\rvm_{j^\star\rightarrow t}.
\label{eq:update_context}
\end{equation}
After the update, we recompute $g_j$ for the remaining candidates using the updated accumulated coverage map $\rvm^{\mathrm{ctx}}_t$ and repeat this process until $|\mathcal{H}_t|=n$.
By selecting frames according to residual coverage gain, \modelname explicitly discourages redundancy among retrieved memories.
Each selected frame is encouraged to cover target-view regions that are not already covered by the current context or by previously selected frames.
The resulting set provides collectively informative geometric evidence by prioritizing complementary target-view coverage.
For multi-frame generation, we apply the greedy selection using the sum of residual coverage gains across target views.
Algorithm~\ref{alg:coverage} summarizes the procedure for retrieving $n$ historical frames for generating the target frame $t$.
An analysis of the greedy approximation and a comparison with exhaustive selection are provided in Appendix~\ref{sec:greedy_analysis}.

\subsection{Scalable depth estimation via sliding-window caching}
\label{sec:geometry_estimation}
We assume access to a per-frame depth map $\mathbf{d}_i$ for each source frame in Eq.~\ref{eq:coverage_map}. We obtain these depths using a pretrained multi-view geometry model, VGGT~\cite{wang2025vggt}, which predicts per-frame depth and relative camera motion from a set of input views.
However, repeatedly applying VGGT to the entire generated history becomes increasingly costly as the rollout grows.
To make depth-based coverage estimation scalable, we introduce a \emph{sliding-window depth caching} strategy.

Let $L$ denote the window size.
After generating frame $\mathbf{x}_t^0$, we run VGGT on $\mathbf{x}_{\mathcal{W}_t}^0$, where $\mathcal{W}_t=\{t-L+1,\ldots,t\}$, and cache the depth maps of newly generated frames for subsequent retrieval.
Because consecutive windows overlap, previously generated frames are included together with newly generated frames, providing local multi-view context for depth estimation across adjacent steps.
When a frame $j$ is added to the memory bank, its depth map $\rvd_j$ is stored together with its RGB frame $\rvx^0_j$ and camera pose $\rvp_j$.
Later, for a target frame $\tau>t$, \modelname computes $\rvm_{j\rightarrow\tau}$ using the cached depth $\rvd_j$ and target pose $\rvp_\tau$.

This strategy keeps geometry estimation efficient during long-horizon generation.
VGGT is evaluated only on a fixed-size local window, so the per-step geometry estimation cost remains bounded by $L$, rather than growing with the length of the generated video.
The window determines which frames are processed together by VGGT, while historical frames retained in the memory bank remain available for retrieval even after leaving this window.
Their cached depth maps are reused for target-view coverage computation throughout the autoregressive rollout, without repeatedly re-estimating geometry for the full memory bank.
\section{Experiments}
\textbf{Datasets.}
\begin{table}[t]
\centering
\caption{\textbf{Consistency evaluation on RealEstate10K and DL3DV10K.}
We report long-horizon consistency and generation quality on RealEstate10K and DL3DV10K.
Lower is better for MEt3R, LPIPS, FVD, and FID, while higher is better for PSNR and SSIM.}
\label{tab:main}
\definecolor{oursbg}{RGB}{234,242,248}

\vspace{2pt}
\setlength{\tabcolsep}{5.2pt}
\renewcommand{\arraystretch}{1.12}

\resizebox{\linewidth}{!}{%
\begin{tabular}{llcccccc}
\toprule
\multirow{2}{*}{\textbf{Dataset}} &
\multirow{2}{*}{\textbf{Method}} &
\multicolumn{4}{c}{\textbf{Consistency}} &
\multicolumn{2}{c}{\textbf{Quality}} \\
\cmidrule(lr){3-6}
\cmidrule(lr){7-8}
& &
\textbf{MEt3R} $\downarrow$ &
\textbf{LPIPS} $\downarrow$ &
\textbf{PSNR} $\uparrow$ &
\textbf{SSIM} $\uparrow$ &
\textbf{FVD} $\downarrow$ &
\textbf{FID} $\downarrow$ \\
\midrule

\multirow{4}{*}{\makecell[l]{RealEstate10K\\[-1pt]\scriptsize $(T{=}40)$}}
& DFoT (No memory-augmentation)
& 0.415 & 0.588 & 11.7 & 0.372 & 265.03 & 29.88 \\
\cmidrule(lr){2-8}
& + WorldMem
& 0.144 & 0.197 & 18.2 & 0.555 & 242.36 & 24.77 \\
& + VMem
& 0.120 & 0.174 & 18.6 & 0.568 & 231.77 & 24.05 \\
& + Spatia
& 0.140 & 0.204 & 17.9 & 0.575 & 228.68 & \textbf{22.60} \\
\rowcolor{oursbg}
\cellcolor{white}
& + \textbf{\modelname (Ours)}
& \textbf{0.102} & \textbf{0.154} & \textbf{18.9} & \textbf{0.576} & \textbf{226.40} & 23.00 \\

\midrule

\multirow{4}{*}{\makecell[l]{DL3DV10K\\[-1pt]\scriptsize $(T{=}100)$}}
& DFoT (No memory-augmentation)
& 0.430 & 0.529 & 14.5 & 0.394 & 706.70 & 88.01 \\
\cmidrule(lr){2-8}
& + WorldMem
& 0.316 & 0.401 & 16.2 & 0.436 & 428.49 & 58.24 \\
& + VMem
& 0.315 & 0.380 & 16.3 & 0.447 & 394.79 & 57.61 \\
& + Spatia
& 0.327 & 0.416 & 15.9 & 0.434 & 439.64 & 56.10 \\
\rowcolor{oursbg}
\cellcolor{white}
& + \textbf{\modelname (Ours)}
& \textbf{0.224} & \textbf{0.308} & \textbf{17.9} & \textbf{0.504} & \textbf{321.81} & \textbf{46.92} \\

\bottomrule
\end{tabular}%
}
\vspace{-6mm}
\end{table}
We evaluate \modelname on RealEstate10K~\cite{zhou2018stereo} and DL3DV10K~\cite{ling2024dl3dv}.
We additionally evaluate dynamic scenes with WorldPlay-8B~\cite{sun2026worldplay} on the SpatialVID dataset~\cite{wang2026spatialvid}.
For RealEstate10K, we construct a custom loop-closing evaluation protocol to explicitly test long-horizon revisitation consistency, where the camera trajectory returns to its initial pose after a long temporal gap.
For DL3DV10K, we use the original trajectories, as the dataset already contains diverse scenes and camera motions that explore larger spatial extents over longer horizons.
Details of dataset construction and trajectory visualizations are provided in the supplementary material.

\textbf{Implementation details.}
For the main comparisons, we use the Diffusion Forcing Transformer (DFoT)~\cite{song2025historyguidedvideodiffusion} as the video generation backbone. 
For RealEstate10K~\cite{zhou2018stereo}, we initialize from the DFoT checkpoint pretrained on RealEstate10K and attach a memory-attention module following WorldMem~\cite{xiao2025worldmem}.
We freeze the DFoT backbone and train only the memory conditioning module for 50K steps on 8 H100 GPUs with a batch size of 64.
For DL3DV10K~\cite{ling2024dl3dv}, we first adapt the RealEstate10K pretrained DFoT by finetuning it on DL3DV10K for 80K steps on 8 GPUs with a batch size of 48.
We then train the memory-augmented model on DL3DV10K with full finetuning for 80K steps with a batch size of 32.
At inference time, we retrieve $n=2$ memory frames at each generation step. We use a temporal context window of $\delta=5$ for RealEstate10K and $\delta=2$ for DL3DV10K.
Additional training and evaluation details are provided in the supplementary material.

\textbf{Baselines.}
We compare \modelname against four baselines:
(1) \textbf{DFoT}~\cite{song2025historyguidedvideodiffusion}, an autoregressive diffusion model that conditions generation on a fixed-length temporal context window; while strong over short horizons, it has no access to frames beyond the window and therefore often degrades when revisiting earlier views.
(2) \textbf{WorldMem}~\cite{xiao2025worldmem}, which augments DFoT with a memory bank and retrieves additional historical frames based on field-of-view (FoV) overlap.
(3) \textbf{VMem}~\cite{li2025vmem}, which retrieves historical views from an explicit surfel-indexed scene memory and optimizes newly predicted depth maps for geometric alignment during memory updates.
(4) \textbf{Spatia}~\cite{zhao2026spatia}, which constructs and updates a 3D scene point cloud as persistent spatial memory, and retrieves reference frames based on 3D IoU between registered, view-specific point clouds.
For the comparisons, we evaluate the WorldMem, VMem, Spatia, and \modelname{} retrieval modules using the same DFoT backbone, memory-conditioning architecture, and checkpoint.

\textbf{Metrics.}
To quantify long-horizon consistency, we report LPIPS~\cite{zhang2018lpips}, PSNR, and SSIM~\cite{WangBSS04ssim} using the dataset-specific evaluation protocols described in Appendix~\ref{sec:protocol}.
To evaluate geometric consistency, we use MEt3R~\cite{asim24met3r}, which estimates pixel correspondences using DUSt3R~\cite{dust3r_cvpr24} and compares matched features from a pretrained image encoder such as DINO~\cite{caron2021emerging}.
In addition, we report FVD~\cite{unterthiner2018fvd} and FID~\cite{heuselRUNH17fid} to evaluate the visual quality of the generated videos.
Together, these metrics provide a comprehensive evaluation of visual quality and long-horizon geometric consistency.
\subsection{Geometric consistency evaluation}
\label{sec:geometric_consistency}
\begin{figure*}[t]
    \centering
    \includegraphics[width=\textwidth]{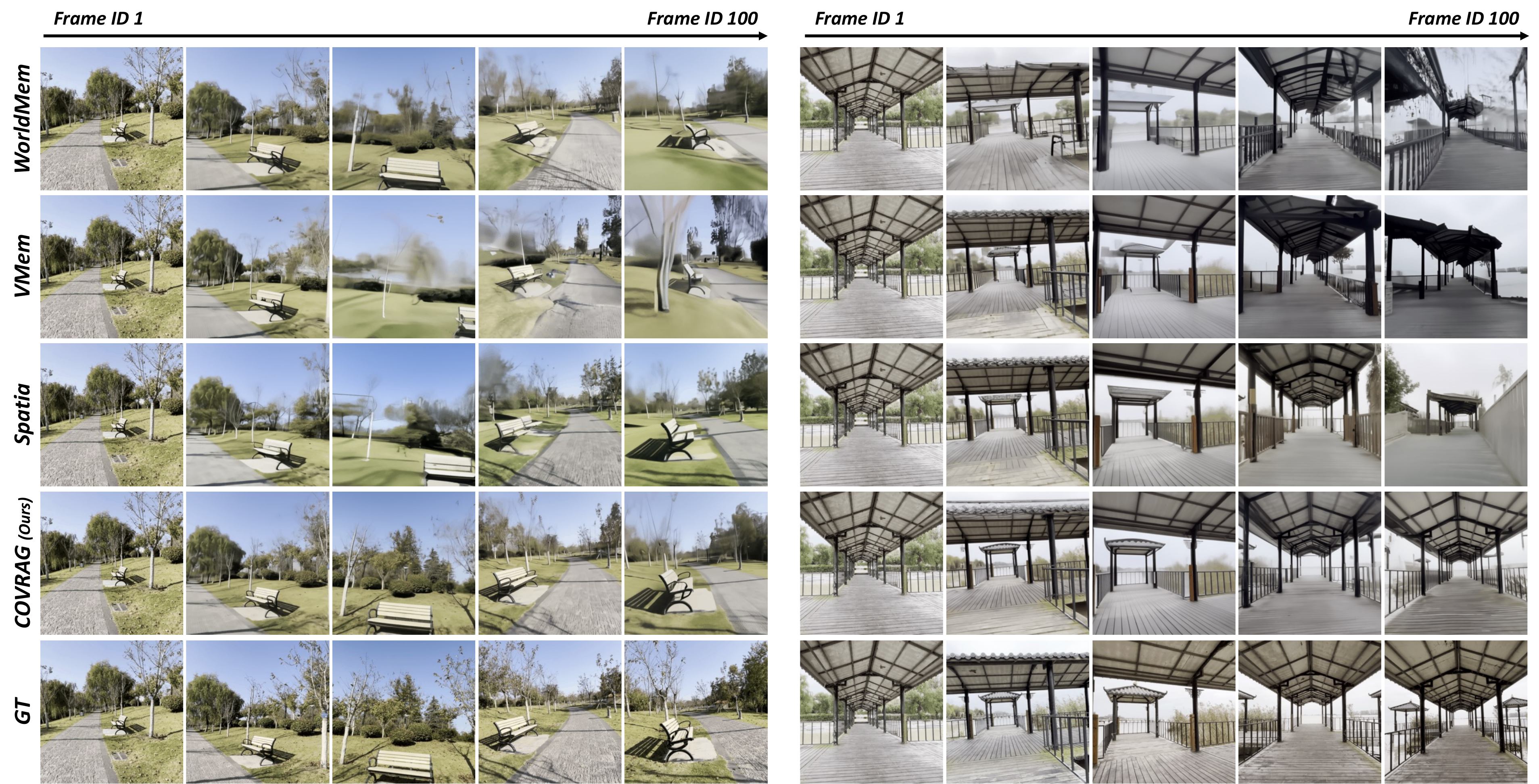}
    \caption{\textbf{Qualitative comparison on DL3DV10K.} We compare 100-frame autoregressive generation using different memory-retrieval modules on a common DFoT backbone, with ground-truth video.}
\label{fig:dl3dv_main_qual}
\vspace{-6mm}
\end{figure*}

Table~\ref{tab:main} reports consistency and generation quality on RealEstate10K~\cite{zhou2018stereo} and DL3DV10K~\cite{ling2024dl3dv}, using a loop-closing evaluation for RealEstate10K and comparisons with ground-truth frames for DL3DV10K.

\textbf{RealEstate10K.}
We construct a loop-closing trajectory by reordering the original camera trajectory so that the sequence approximately returns to the initial view.
This setting directly tests whether the model can preserve geometric consistency when revisiting regions observed only at the beginning of the rollout.
As shown in Table~\ref{tab:main}, DFoT performs poorly in this revisitation setting, achieving 0.415 MEt3R and 0.588 LPIPS.
This indicates that a fixed temporal context is insufficient for maintaining long-term geometric consistency in this setting.
Memory-augmented methods substantially improve consistency, demonstrating the importance of accessing historical frames beyond the local context window.
Among them, \modelname achieves the best performance across all consistency metrics and the lowest FVD.
Compared with WorldMem~\cite{xiao2025worldmem}, which uses FoV-based retrieval, \modelname reduces MEt3R by 29\% $(0.144 \rightarrow 0.102)$ and LPIPS by 22\% $(0.197 \rightarrow 0.154)$, while also improving PSNR $(18.2 \rightarrow 18.9)$ and SSIM $(0.555 \rightarrow 0.576)$.
Compared with VMem~\cite{li2025vmem}, \modelname further improves MEt3R $(0.120 \rightarrow 0.102)$ and LPIPS $(0.174 \rightarrow 0.154)$, without maintaining a persistent global 3D scene representation.
Compared with Spatia~\cite{zhao2026spatia}, \modelname reduces MEt3R by 27\% $(0.140 \rightarrow 0.102)$ and LPIPS by 25\% $(0.204 \rightarrow 0.154)$, while improving PSNR by 1.0\,dB $(17.9 \rightarrow 18.9)$.
In addition, \modelname achieves strong generation quality, with an FVD of 226.40 and an FID of 23.00.
These results support residual target-view coverage as an effective criterion for selecting complementary historical observations under a fixed retrieval budget.

\textbf{DL3DV10K.}
We further evaluate on DL3DV10K to test long-horizon generation in more diverse and spatially extended scenes.
Unlike RealEstate10K, we do not construct artificial loop-closing trajectories.
Instead, we use the original trajectories, which naturally contain diverse camera motions and explore larger spaces over longer horizons.
The DL3DV10K results also show a substantial gap between \modelname and prior retrieval-augmented baselines.
Compared with WorldMem, \modelname reduces MEt3R by 29\% $(0.316 \rightarrow 0.224)$ and LPIPS by 23\% $(0.401 \rightarrow 0.308)$.
It also substantially improves generation quality, reducing FVD by 25\% $(428.49 \rightarrow 321.81)$ and FID by 19\% $(58.24 \rightarrow 46.92)$.
Compared with VMem, which uses explicit geometric memory, \modelname reduces MEt3R by 29\%, LPIPS by 19\%, FVD by 19\%, and FID by 19\%.
Compared with Spatia~\cite{zhao2026spatia}, \modelname also improves MEt3R $(0.327 \rightarrow 0.224)$ and LPIPS $(0.416 \rightarrow 0.308)$, while reducing FVD by 27\% $(439.64 \rightarrow 321.81)$ and FID by 16\% $(56.10 \rightarrow 46.92)$.
These gains suggest that target-view coverage is also effective in diverse 3D scenes with camera trajectories that span larger spatial extents.

\textbf{Qualitative results.}
Figure~\ref{fig:dl3dv_main_qual} compares 100-frame autoregressive generations on DL3DV10K using the WorldMem, VMem, Spatia, and \modelname retrieval modules on a common DFoT backbone.
Compared with the other retrieval variants, \modelname better preserves scene geometry and spatial layout over long camera trajectories.
In the outdoor park sequence, baselines exhibit drift and distortion around the bench, path, and trees, whereas our method maintains sharper object structure and more stable scene composition.
In the covered walkway sequence, baselines progressively deform the roof, pillars, and railings, whereas our method preserves geometry that is more coherent with the ground truth.
These results indicate that coverage-based retrieval provides more informative memory evidence for long-horizon generation in diverse 3D scenes.
Additional qualitative results on RealEstate10K and DL3DV10K are provided in Appendix~\ref{sec:appendix_qual}.
\subsection{Evaluation in dynamic scenes}
\label{sec:dynamic_scenes}

To evaluate retrieval in the presence of scene motion, we integrate \modelname into the pretrained video world model WorldPlay-8B~\cite{sun2026worldplay}. We replace its FoV-based memory-selection rule while keeping the model weights, generation settings, context, and retrieval budget fixed, without retraining.
We evaluate 100 long-video rollouts initialized from 100 SpatialVID~\cite{wang2026spatialvid} source videos selected for dynamic content.
Their mean annotated per-video motion score and dynamic ratio are 4.81 and 10.88\%, respectively.
Each rollout comprises 29 autoregressive chunks (116 latent frames), decoded into 461 video frames including the initial conditioning frame, i.e., 460 newly generated frames. \modelname retrieves memory at every chunk using a 10-frame VGGT inference window. We compute PSNR, SSIM, and LPIPS between generated images at three revisited camera-pose pairs per video (300 pairs in total), and evaluate FVD over the 100 generated long videos.

As shown in Table~\ref{tab:worldplay_dynamic}, \modelname improves PSNR from 14.3 to 14.9 and SSIM from 0.390 to 0.411, while reducing LPIPS from 0.373 to 0.333 and FVD from 418.44 to 376.83.
These results demonstrate that \modelname can serve as a plug-in
retrieval module for a pretrained video world model, improving
pose-revisit metrics and FVD in the evaluated scenes containing
motion without retraining the generator.
\modelname does not explicitly model object dynamics.
Large or rapid object motion can make a geometrically relevant historical frame contain stale object states, limiting the reliability of coverage-based retrieval.
Additional VBench scores~\cite{huang2023vbench}, evaluated over the same 100 generated videos, are reported in Appendix Table~\ref{tab:worldplay_vbench}.
\begin{table}[t]
\definecolor{oursbg}{RGB}{234,242,248}
\centering
\begin{minipage}[t]{0.48\linewidth}
    \vspace{0pt}
    \centering
    \caption{\textbf{Evaluation in dynamic scenes.}
    WorldPlay-8B on the SpatialVID dynamic-scene subset.
    Its FoV-based memory selection is replaced by COVRAG
    without retraining.}
    \label{tab:worldplay_dynamic}
    \setlength{\tabcolsep}{2pt}
    \begin{tabular}{@{}lrrrr@{}}
    \toprule
    Method & PSNR$\uparrow$ & SSIM$\uparrow$ & LPIPS$\downarrow$ & FVD$\downarrow$ \\
    \midrule
    WorldPlay-8B & 14.3 & 0.390 & 0.373 & 418.44 \\
    \rowcolor{oursbg}
    + \textbf{\modelname}
        & \textbf{14.9} & \textbf{0.411}
        & \textbf{0.333} & \textbf{376.83} \\
      \bottomrule
    \end{tabular}
  \end{minipage}\hfill
  \begin{minipage}[t]{0.48\linewidth}
    \vspace{0pt}
    \centering
    \caption{\textbf{Complementarity with 3D conditioning.}
    GEN3C on RealEstate10K, with its reconstruction and rendering
    pipeline unchanged. No retraining is used.}
    \label{tab:gen3c_complementarity}
    \setlength{\tabcolsep}{2pt}
    \begin{tabular}{@{}lrrrr@{}}
      \toprule
      Method & PSNR$\uparrow$ & SSIM$\uparrow$
             & LPIPS$\downarrow$ & MEt3R$\downarrow$ \\
      \midrule
      GEN3C
        & 14.2 & 0.572 & 0.519 & 0.377 \\
        \rowcolor{oursbg}
      + \textbf{\modelname}
        & \textbf{17.6} & \textbf{0.675}
        & \textbf{0.349} & \textbf{0.251} \\
      \bottomrule
    \end{tabular}
  \end{minipage}
\vspace{-6mm}
\end{table}

\subsection{Complementarity with 3D conditioning}
\label{sec:complementarity}
GEN3C~\cite{ren2025gen3c} renders point clouds for conditioning, allowing us to test whether complementary reference selection further improves consistency.
We integrate \modelname into its official long-rollout pipeline with the pretrained Cosmos-7B world model~\cite{nvidia2025cosmosworldfoundationmodel}, without retraining.
After the initial chunk, \modelname selects GEN3C's two reference frames while retaining its conditioning mechanism, point-cloud budget, reconstruction and rendering procedures, and generation settings.
We use identical camera trajectories and random seeds for the two variants. 
We evaluate 100 RealEstate10K videos of 361 frames each, measuring PSNR, SSIM, LPIPS, and MEt3R between generated images at 12 revisited camera-pose pairs per video (1,200 pairs in total).

Table~\ref{tab:gen3c_complementarity} shows that \modelname increases PSNR from 14.2 to 17.6 and SSIM from 0.572 to 0.675, while reducing LPIPS from 0.519 to 0.349 and MEt3R from 0.377 to 0.251.
These improvements show that coverage-based reference selection complements GEN3C's reconstruction-and-rendering pipeline while preserving its conditioning mechanism and point-cloud budget.
On the same hardware, coverage computation and selection require 5.8\,s, or approximately 1.3\% of GEN3C's 7.6\,min runtime.
\subsection{Analysis}
\label{sec:ablation}
\begin{figure}[t]
\centering

\begin{minipage}[t][5.4cm][s]{0.48\linewidth}
\vspace{0pt}
\centering

\captionof{table}{
    \textbf{Ablation of \modelname on RE10K.}
    FoV and Coverage denote FoV overlap and target-view
    coverage, respectively.
    \texttt{Indep.} and \texttt{Res.} denote independent
    and residual gain.
}
\label{tab:ablation}

\par\vfill

\setlength{\tabcolsep}{3pt}
\renewcommand{\arraystretch}{1.15}

\begin{tabular*}{\linewidth}{@{\extracolsep{\fill}}lllcc@{}}
\toprule
\multirow{2}{*}{\textbf{ID}}
    & \multicolumn{2}{c}{\textbf{Retrieval}}
    & \multicolumn{2}{c}{\textbf{Consistency}} \\
\cmidrule(lr){2-3}
\cmidrule(lr){4-5}
    & \textbf{Evidence}
    & \textbf{Selection}
    & \textbf{MEt3R} $\downarrow$
    & \textbf{LPIPS} $\downarrow$ \\
\midrule
(a) & FoV      & \texttt{Indep.} & 0.141 & 0.198 \\
(b) & FoV      & \texttt{Res.}   & 0.130 & 0.187 \\
(c) & Coverage & \texttt{Indep.} & 0.149 & 0.210 \\
\midrule
(d) & Coverage & \texttt{Res.}
    & \textbf{0.100} & \textbf{0.156} \\
\bottomrule
\end{tabular*}

\par\vspace{0pt}
\end{minipage}%
\hfill
\begin{minipage}[t][5.4cm][s]{0.5\linewidth}
\vspace{0pt}
\centering

\includegraphics[width=\linewidth]{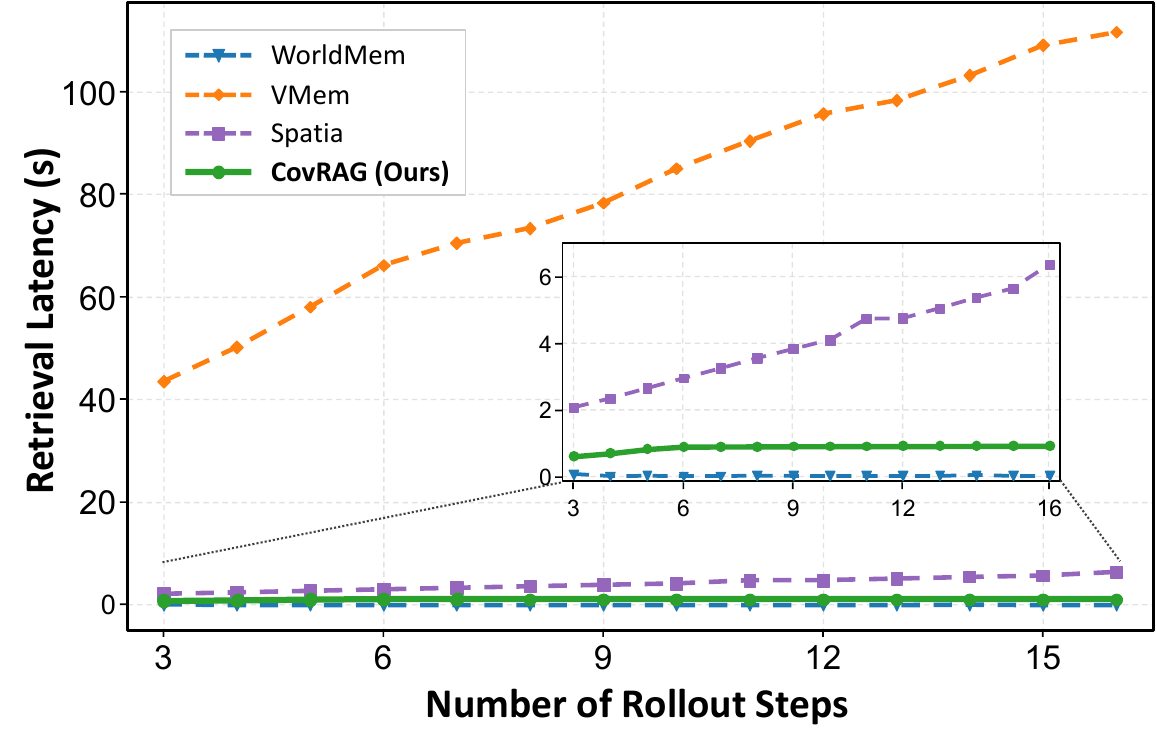}

\par\vfill

\captionof{figure}{
    Retrieval latency comparison during autoregressive rollout.
}
\label{fig:latency_fig}

\par\vspace{0pt}
\end{minipage}
\vspace{-2mm}
\end{figure}

\vspace{-1mm}
\textbf{Ablation study.}
We ablate two key design choices of \modelname: \emph{what} 3D-aware evidence is used for retrieval and \emph{how} frames are selected from that evidence.
For evidence, we compare coarse field-of-view (FoV) overlap with our depth-based target-view coverage map.
For selection, we compare independent gain, which ranks candidates independently, with residual gain, which iteratively selects frames that cover currently unsupported target-view regions.
Table~\ref{tab:ablation} shows that combining target-view coverage with residual-gain selection yields the best results.
With FoV evidence, replacing independent gain with residual gain improves MEt3R from $0.141$ to $0.130$ and LPIPS from $0.198$ to $0.187$, indicating that redundancy-aware selection is useful even with coarse view-level evidence.
However, the gains are limited because FoV overlap cannot localize which target pixels are geometrically supported.
Conversely, using target-view coverage with independent selection performs poorly, yielding $0.149$ MEt3R and $0.210$ LPIPS.
This indicates that fine-grained coverage evidence alone is insufficient without a selection rule that accounts for complementary coverage.
Combining target-view coverage with residual gain achieves the best results, reducing MEt3R to $0.100$ and LPIPS to $0.156$.
These results support the central design principle of \modelname.
We further examine sensitivity to estimated geometry through controlled perturbations and longer autoregressive rollouts in Appendix~\ref{sec:vggt_dependency}.

\textbf{Retrieval cost comparison.}
Figure~\ref{fig:latency_fig} compares the retrieval latency of different memory-augmented generation methods during autoregressive rollout.
VMem's retrieval incurs substantially higher average latency, which increases with rollout length, since the explicit 3D memory becomes more expensive to update and query over longer rollouts.
Spatia's retrieval is substantially faster than VMem, but its latency also increases with rollout length and remains higher than that of \modelname.
In contrast, WorldMem achieves the lowest latency by using lightweight pose-level FoV overlap for retrieval.
However, this efficiency comes at the cost of relying on coarse view-level evidence, which limits long-term geometric consistency, as shown in Table~\ref{tab:main}.
Although \modelname performs depth-based target-view coverage computation, including VGGT inference, warping historical frames into the target view, and selecting frames by residual coverage gain, its per-step retrieval latency remains nearly constant as the rollout grows.
This is enabled by sliding-window depth caching: depth is estimated only within a fixed local window, and the resulting depth maps are reused for subsequent coverage computation.
Consequently, \modelname provides pixel-level geometric evidence for retrieval without maintaining and updating a persistent global 3D scene representation.
These results show that depth-based target-view coverage provides a practical middle ground between coarse but efficient FoV-based retrieval and fine-grained but costly explicit 3D memory.

\vspace{-2mm}
\section{Conclusion}
\vspace{-2mm}
We presented \modelname, a depth-based memory-retrieval framework to improve long-term geometric consistency in autoregressive video generation.
\modelname constructs target-view coverage maps as lightweight 3D-aware evidence and selects memory frames by residual coverage gain, encouraging retrieved frames to provide complementary support for uncovered target-view regions.
With sliding-window depth caching, \modelname enables pixel-level coverage reasoning without maintaining an explicit global 3D reconstruction.
Experiments on RealEstate10K and DL3DV10K show that target-view coverage is an effective retrieval objective for 3D-consistent long-horizon generation.

\clearpage
\begin{ack}
This work was partly supported by the Institute of Information \& Communications Technology Planning \& Evaluation (IITP) grants funded by the Korea government (MSIT) (No. RS-2025-25442149, LG AI STAR Talent Development Program for Leading Large-Scale Generative AI Models in the Physical AI Domain, 30\%; No. RS-2019-II190075, Artificial Intelligence Graduate School Program (KAIST), 10\%) and by the InnoCORE program of the Ministry of Science and ICT (AI Meta-Scientist, N10260110, 30\%). This work was also supported by the National Supercomputing Center with supercomputing resources including technical support (KSC-2025-CRE-0506, 30\%). The authors declare no competing interests.
\end{ack}
{
\bibliography{main}
\bibliographystyle{unsrt}
}

\appendix

\newpage

\section{Experimental Details}
\subsection{Baseline Setup}
We provide additional details on the baseline configurations used in the main RealEstate10K and DL3DV10K comparisons.
Within each dataset, WorldMem, VMem, Spatia, and \modelname share the same trained DFoT~\cite{song2025historyguidedvideodiffusion} checkpoint, including the memory-conditioning weights.
These variants differ in the retrieval module used to select historical frames.
The DFoT baseline uses temporal context without memory augmentation, as specified below.

\paragraph{DFoT.} 
DFoT~\cite{song2025historyguidedvideodiffusion} conditions generation on a fixed-length temporal context window without memory retrieval.
For RealEstate10K, we use seven context frames.
For DL3DV10K, we use four context frames and generate four target frames per autoregressive step.
These settings provide conditioning budgets comparable to those of the retrieval-augmented variants.

\paragraph{WorldMem.} 
WorldMem~\cite{xiao2025worldmem} retrieves historical frames using pose-based field-of-view (FoV) overlap.
For RealEstate10K, we use five recent context frames and retrieve two memory frames at each generation step.
For DL3DV10K, we use two context frames, retrieve two memory frames, and generate four target frames per autoregressive step.
WorldMem, VMem, Spatia, and \modelname use these same context, target, and memory-retrieval settings.
The retrieval module and its hyperparameters follow the official implementation.

\paragraph{VMem.} 
VMem~\cite{li2025vmem} uses CUT3R~\cite{cut3r} to estimate geometry for its explicit surfel-indexed scene memory and SEVA~\cite{zhou2025stable} as its novel-view generator.
During memory updates, it optimizes newly predicted depth maps for geometric alignment.
It renders view-index attributes stored in the surfel memory from the target viewpoint to identify relevant historical frames.
In our comparison, we adapt its retrieval module to the shared DFoT checkpoint and memory-conditioning module, retrieving two RGB memory frames under the same conditioning settings specified above.
All retrieval hyperparameters follow the default settings provided in the official implementation.

\paragraph{Spatia.} 
Spatia~\cite{zhao2026spatia} maintains a global scene point cloud reconstructed with MapAnything~\cite{keetha2026mapanything} as persistent spatial memory.
We adapt its reference-frame retrieval to the same DFoT checkpoint and memory-conditioning module used by the other baselines, retrieving two RGB memory frames.
To avoid GPU out-of-memory errors during reconstruction on DL3DV10K, we temporally subsample the historical frames used for reconstruction with a mapping stride of $6$.
Our implementation is based on the official Spatia code.

\paragraph{\modelname.}
For both datasets, we use a fixed VGGT~\cite{wang2025vggt} window of 10 frames.
Predicted depth maps are cached and reused for subsequent target-view coverage computation, following the sliding-window depth caching procedure described in the main paper.

\subsection{Model training details}
For RealEstate10K, we initialize from the DFoT checkpoint pretrained on RealEstate10K.
We freeze the DFoT backbone and train only the memory-attention conditioning module, since the pretrained backbone already provides a strong pose-conditioned generation prior on this dataset. 
For DL3DV10K, we first adapt the RealEstate10K pretrained DFoT backbone by fine-tuning it on DL3DV10K.
We then train the memory-augmented model with full fine-tuning.
This is because DL3DV10K contains more diverse scenes and wider camera trajectories, and the retrieved memory frames can have larger pose differences from the current context.
In this setting, training only the memory-attention module did not reliably encourage the model to use the retrieved frames.
During memory-augmented training, we sample memory frames using a pose-based proxy and retain memory dropout from the WorldMem training setup.
We use a temporal buffer to avoid overly distant or irrelevant memories, and mix easy and hard memory samples based on their pose relation to the target view.
We gradually increase the sampling range and context-frame stride during training.
This exposes the model to imperfect or missing historical observations.
Retrieved memories therefore serve as auxiliary evidence alongside the recent temporal context.

\subsection{Evaluation protocol}
\label{sec:protocol}
\begin{figure*}[t]
    \centering
    \includegraphics[width=\columnwidth]{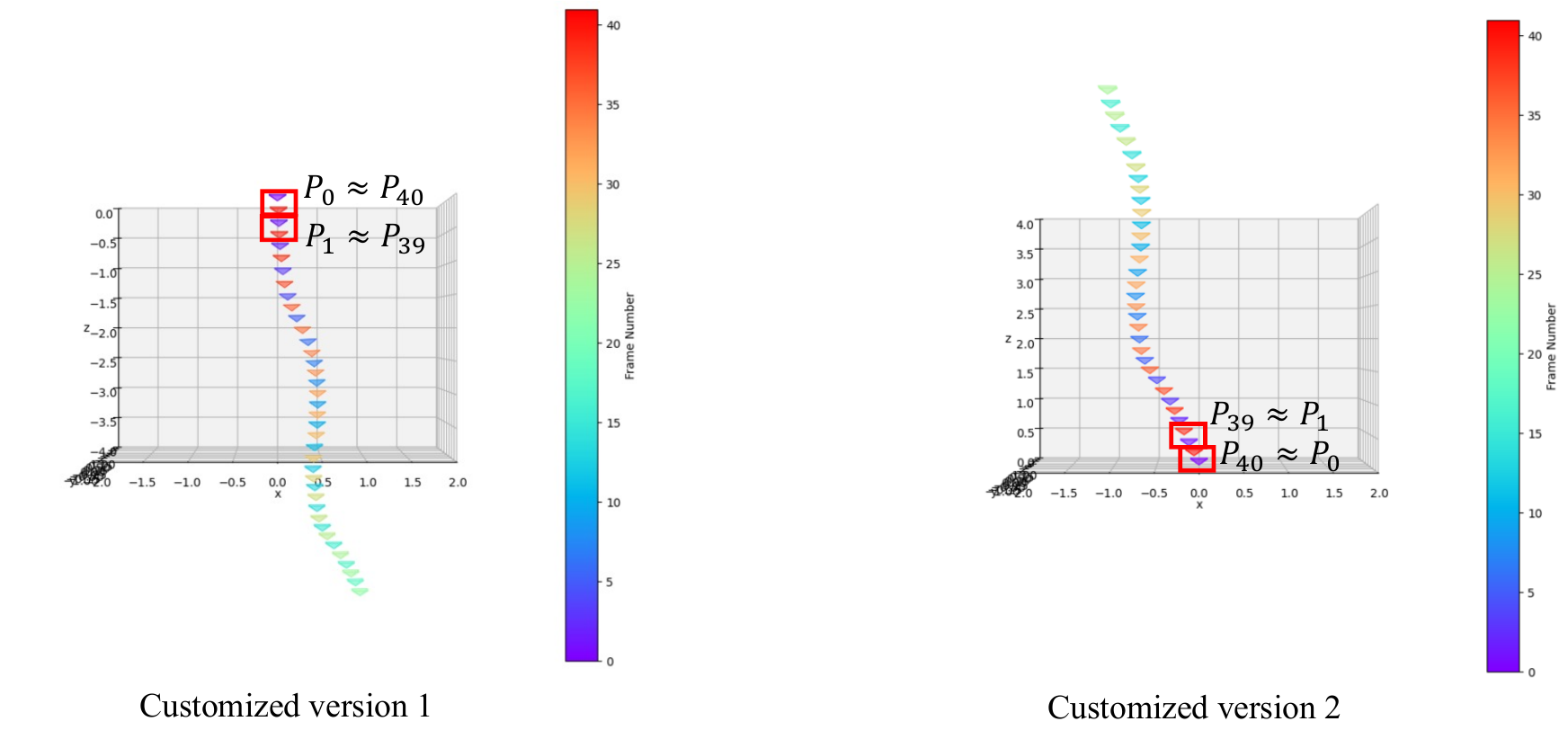}
    \caption{\textbf{Example customized loop-closing camera trajectories for RealEstate10K evaluation.} Starting from 41-frame RealEstate10K clips, we illustrate two example loop-closing variants obtained by reordering camera poses so that late frames revisit poses close to early frames. Here, $P_i$ denotes the camera pose at frame $i$. The constructed loops make the final poses approximately match the initial poses, e.g., $P_0 \approx P_{40}$ and $P_1 \approx P_{39}$. The long temporal gap between the initial observation and the revisited view provides a controlled benchmark for measuring long-term drift and revisitation consistency.}
    \label{fig:pose_custom}
\end{figure*}
\begin{figure*}[t!]
    \centering
    \includegraphics[width=0.9\columnwidth]{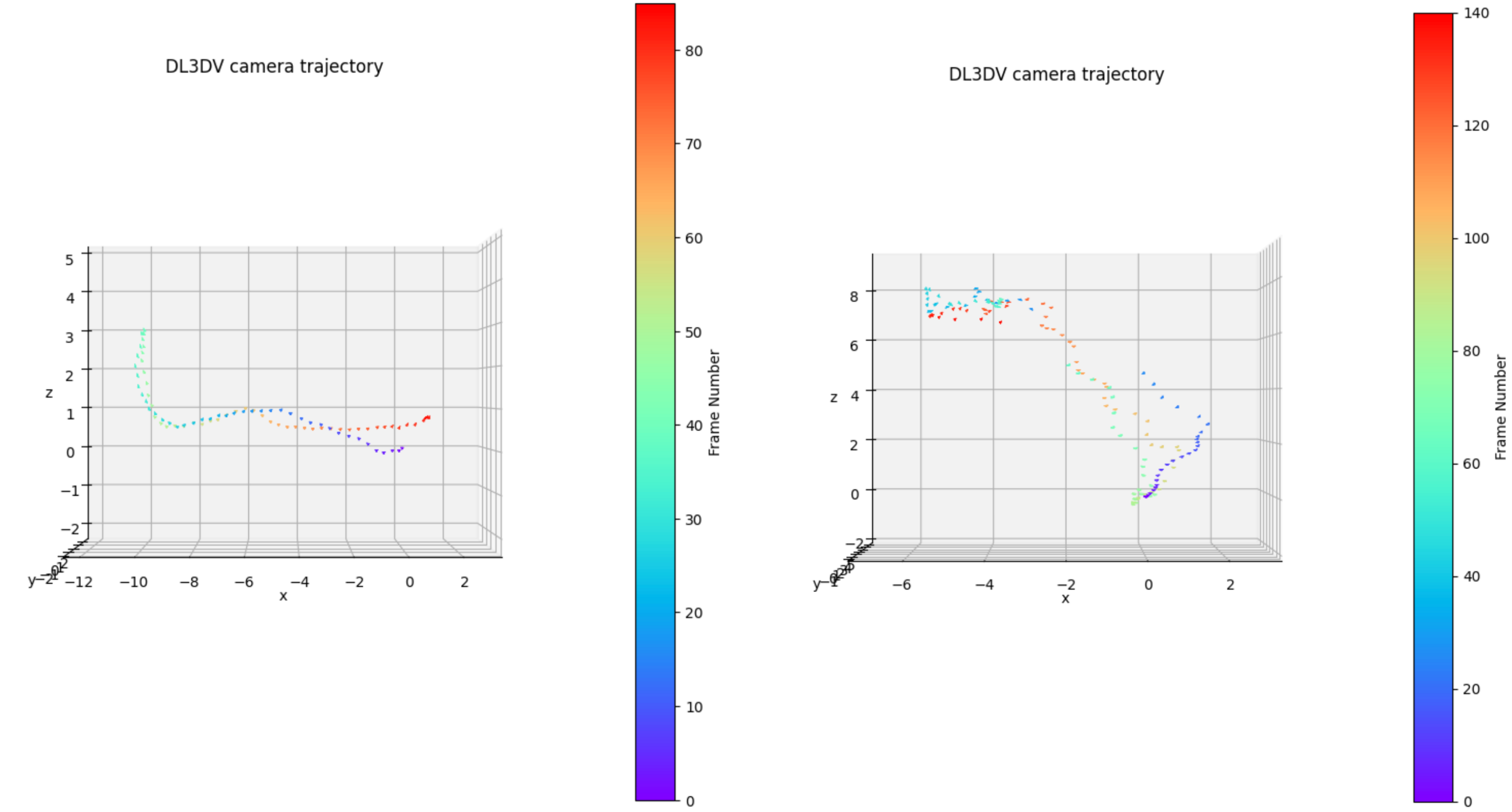}
    \caption{\textbf{Examples of DL3DV camera trajectories.} We show two representative examples of the original camera paths in the DL3DV10K, with colors indicating temporal progression. These trajectories illustrate broad viewpoint changes and partial scene observations, complementing the controlled RealEstate10K loop-closing setting.}
    \label{fig:dl3dv_trajectory}
\end{figure*}
\paragraph{RealEstate10K.} RealEstate10K~\cite{zhou2018stereo} provides real-world video clips with ground-truth camera trajectories.
To evaluate long-term consistency, we adopt a loop-closing evaluation on RealEstate10K.
We intentionally customize the trajectories to force revisitation.
The main quantitative results are reported on 400 test video clips, while analysis and ablation studies are conducted on 200 test video clips.
Specifically, given a clip of 41 frames, we reorder the camera trajectory such that the camera first moves away from the initial viewpoint and then returns to a spatially similar pose after a long temporal gap, as illustrated in Figure~\ref{fig:pose_custom}.
Each rollout starts from a single initial conditioning frame, with subsequent frames generated autoregressively.
This design enables a controlled assessment of long-term consistency: failures in retrieval at intermediate steps can lead the model to condition on diverging context, causing errors to accumulate over time.
As a consequence, discrepancies between the initial frames and the final revisited frames directly reflect long-term drift induced by insufficient retrieval.
Therefore, we compute MEt3R, LPIPS, PSNR, and SSIM on two early--revisited image pairs per video, formed by matching the first two frames of each generated sequence with the corresponding revisited frames at the end of the loop.
These comparisons assess how well scene geometry and appearance are preserved across the temporal gap.

\begin{figure*}[t]
    \centering
    \includegraphics[width=\textwidth]{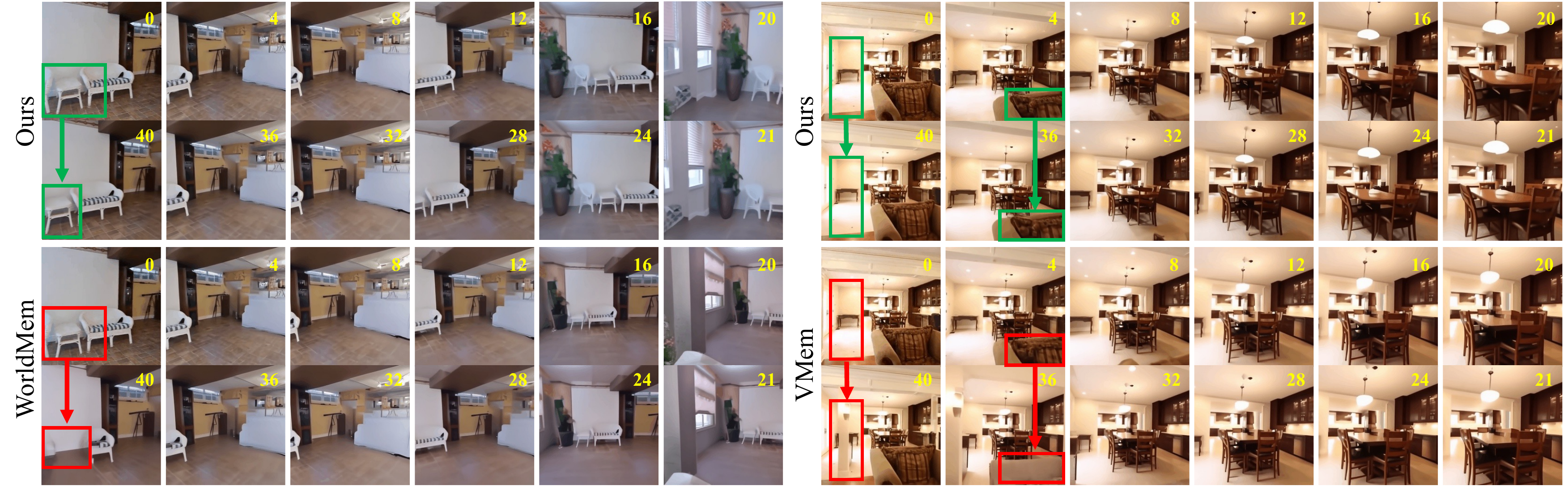}
    \caption{\textbf{Qualitative comparison on loop-closing trajectories.}
We visualize two RealEstate10K sequences generated along loop-closing camera paths, where the camera revisits earlier views after a temporal gap.
\modelname (top row) better preserves scene geometry and appearance across revisitation (highlighted in \textbf{\textcolor{ForestGreen}{green}}), yielding consistent generation of previously observed regions.
In contrast, WorldMem and VMem (bottom row) exhibit long-horizon inconsistencies such as spatial drift and missing or hallucinated structures (highlighted in \textbf{\textcolor{red}{red}}), despite using memory retrieval.
Frame indices denote the generation order along the loop.}
\label{fig:supple_re10k_qual}
\end{figure*}

\begin{figure*}[t]
    \centering
    \includegraphics[width=\columnwidth]{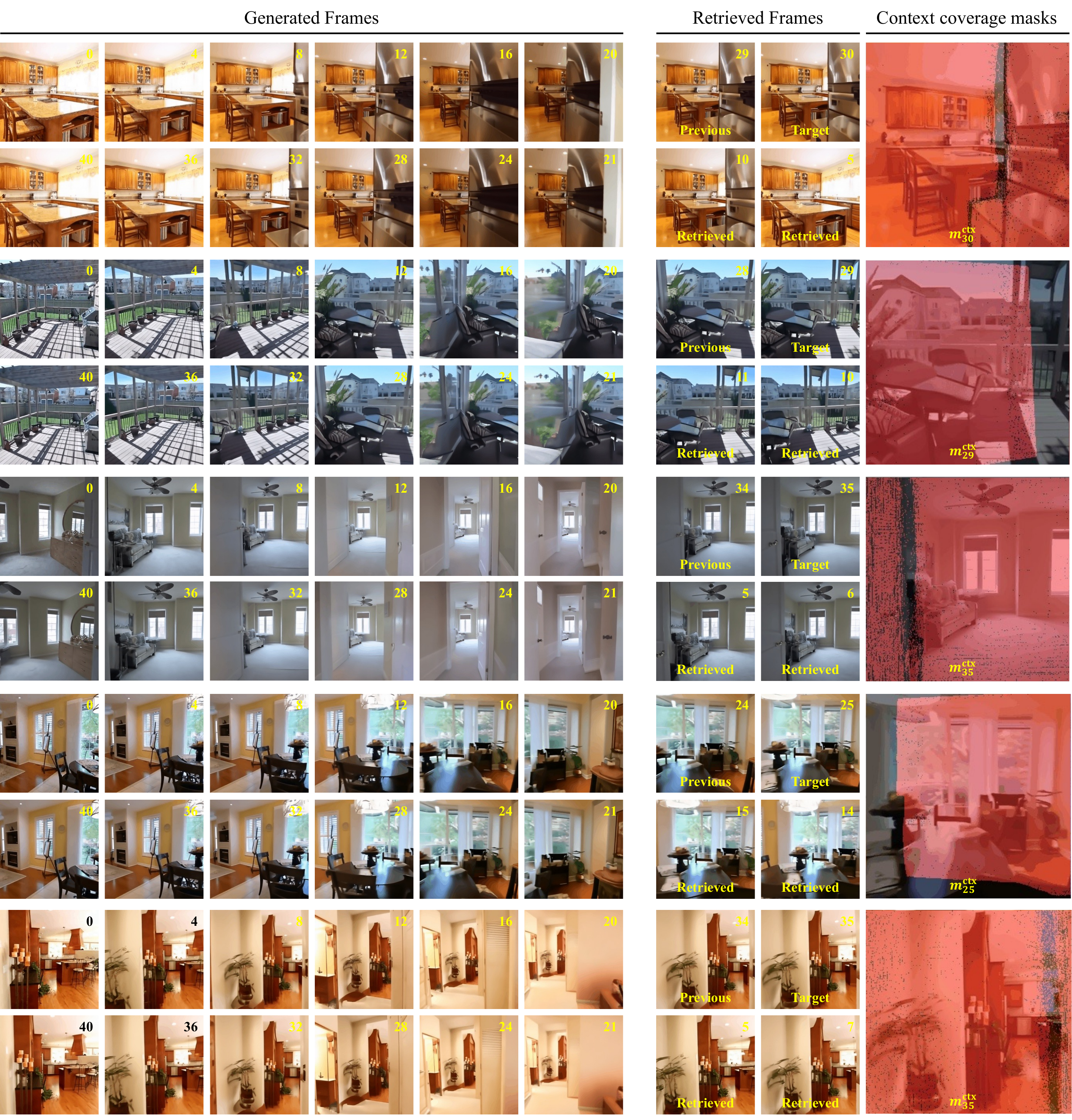}
    \caption{\textbf{Retrieval behavior and context coverage visualizations on RealEstate10K.} For each RealEstate10K loop-closing example, the left panel shows generated frames along the trajectory.
    The middle panel shows the previous and target  frames for reference in its upper row and the two retrieved memory frames in its lower row. 
    The right panel shows the context coverage map $\rvm_t^{\mathrm{ctx}}$ before adding retrieved memories. The coverage map visualizes which target-view regions are already supported by the recent temporal context. \modelname retrieves memory frames that provide complementary evidence for regions not sufficiently covered by the context, illustrating how coverage-maximizing retrieval reduces redundant memory selection.}
    \label{fig:appendix_qual_fig}
\end{figure*}
\paragraph{DL3DV10K.}
We further evaluate on DL3DV10K~\cite{ling2024dl3dv}, which contains more diverse scenes and camera trajectories than RealEstate10K.
We report results on 100 videos sampled from DL3DV10K.
Unlike the RealEstate10K evaluation, we do not construct an artificial loop-closing trajectory for DL3DV10K.
Instead, we use the original trajectories, which naturally explore larger spatial extents and include more varied camera motion.
Each rollout is initialized with the provided context frames, after which generation proceeds autoregressively.
Figure~\ref{fig:dl3dv_trajectory} visualizes representative DL3DV10K camera trajectories, showing that the views often cover a wider range of poses and partial observations.

This setting complements the controlled RealEstate10K loop-closing benchmark.
Since the original trajectories traverse larger and more complex 3D spaces, the current temporal context often covers only part of the target view.
The model must therefore retrieve historical frames that provide complementary partial-view evidence.
DL3DV10K thus provides a natural testbed for evaluating target-view coverage-based retrieval under diverse camera motion.
We compute MEt3R, LPIPS, PSNR, and SSIM between generated frames and their corresponding ground-truth frames.

\subsection{GT Pose Alignment for VGGT-based Geometry Estimation}
\label{sec:vggt_scaled}
VGGT~\cite{wang2025vggt} predicts depth and camera motion in relative scale, whereas our diffusion model is driven by a camera trajectory with metric extrinsics.
Directly using VGGT depth with unscaled metric poses can therefore introduce inconsistencies during 3D re-projection to the target frame.
We estimate a single scale factor from the initial VGGT window to align the ground-truth trajectory with its predicted geometry.
Let $\mathbf{t}^{\text{VGGT}}_i$ be VGGT-predicted translations and $\mathbf{t}^{\text{GT}}_i$ be the metric translations for the initial VGGT input window of length $N$.
After normalizing both trajectories to start from the first frame, we compute
\begin{equation}
s \;=\; \frac{1}{N-1}\sum_{i=2}^{N}
\frac{\|\mathbf{t}^{\text{VGGT}}_i-\mathbf{t}^{\text{VGGT}}_1\|_2}
{\|\mathbf{t}^{\text{GT}}_i-\mathbf{t}^{\text{GT}}_1\|_2}.
\end{equation}
We apply this fixed scale to all ground-truth camera translations
throughout the rollout, using
$\hat{\rvp}_i = [\mathbf{R}_i \mid s\mathbf{t}_i]$
for both source and target poses.
This alignment provides an initial scale reference for depth-based reprojection and coverage computation, although scale variation across VGGT windows may still affect the accuracy of coverage estimates.

\section{Additional Results}

\subsection{Qualitative Results}
\label{sec:appendix_qual}

\paragraph{RealEstate10K.}
We include a qualitative comparison with retrieval-based baselines on RealEstate10K in Figure~\ref{fig:supple_re10k_qual}, which highlights how \modelname more consistently preserves scene geometry and appearance when revisiting previously observed views.
When the camera returns to a previously observed viewpoint, \modelname generates the revisited regions with substantially higher geometric and appearance consistency (green highlights), preserving scene layout and object structure across the long temporal gap. In contrast, WorldMem and VMem often fail to recover locally missing content at revisitation, exhibiting structural inconsistencies such as missing objects or hallucinated geometry (red highlights).

We also present additional qualitative results that visualize the generated videos together with the coverage mask used for retrieval, as shown in Figure~\ref{fig:appendix_qual_fig}.
These results make explicit why certain frames are retrieved and illustrate how coverage-maximizing retrieval guides consistent generation.

\paragraph{DL3DV10K.}
\begin{figure*}[t]
    \centering
    \includegraphics[width=\columnwidth]{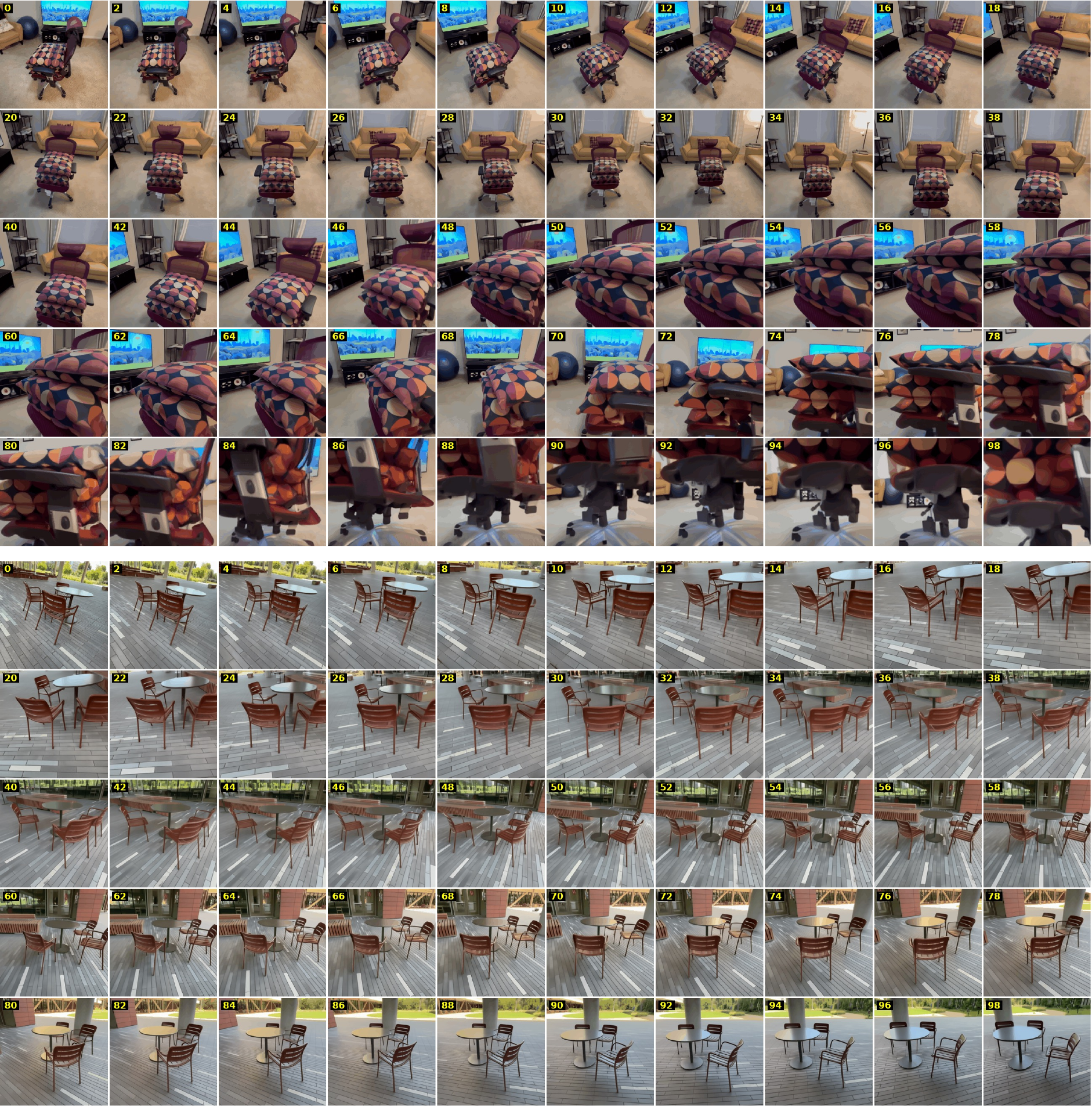}
    \caption{\textbf{Additional qualitative rollouts on DL3DV10K.} We visualize 100-frame autoregressive generations on diverse DL3DV10K scenes and camera trajectories. Frame indices mark temporal order. Across extended rollouts with substantial viewpoint changes, \modelname maintains coherent scene layout and object structure over long horizons, illustrating the qualitative behavior of coverage-based retrieval beyond the controlled RealEstate10K loop-closing setting.}
    \label{fig:dl3dv_supple_qual}
\end{figure*}
We also provide additional qualitative results on DL3DV10K to visualize long-horizon generation under diverse scenes and camera trajectories. 
Figure~\ref{fig:dl3dv_supple_qual} shows 100-frame autoregressive rollout results over extended trajectories.

\begin{figure*}[t]
    \centering
    \includegraphics[width=\columnwidth]{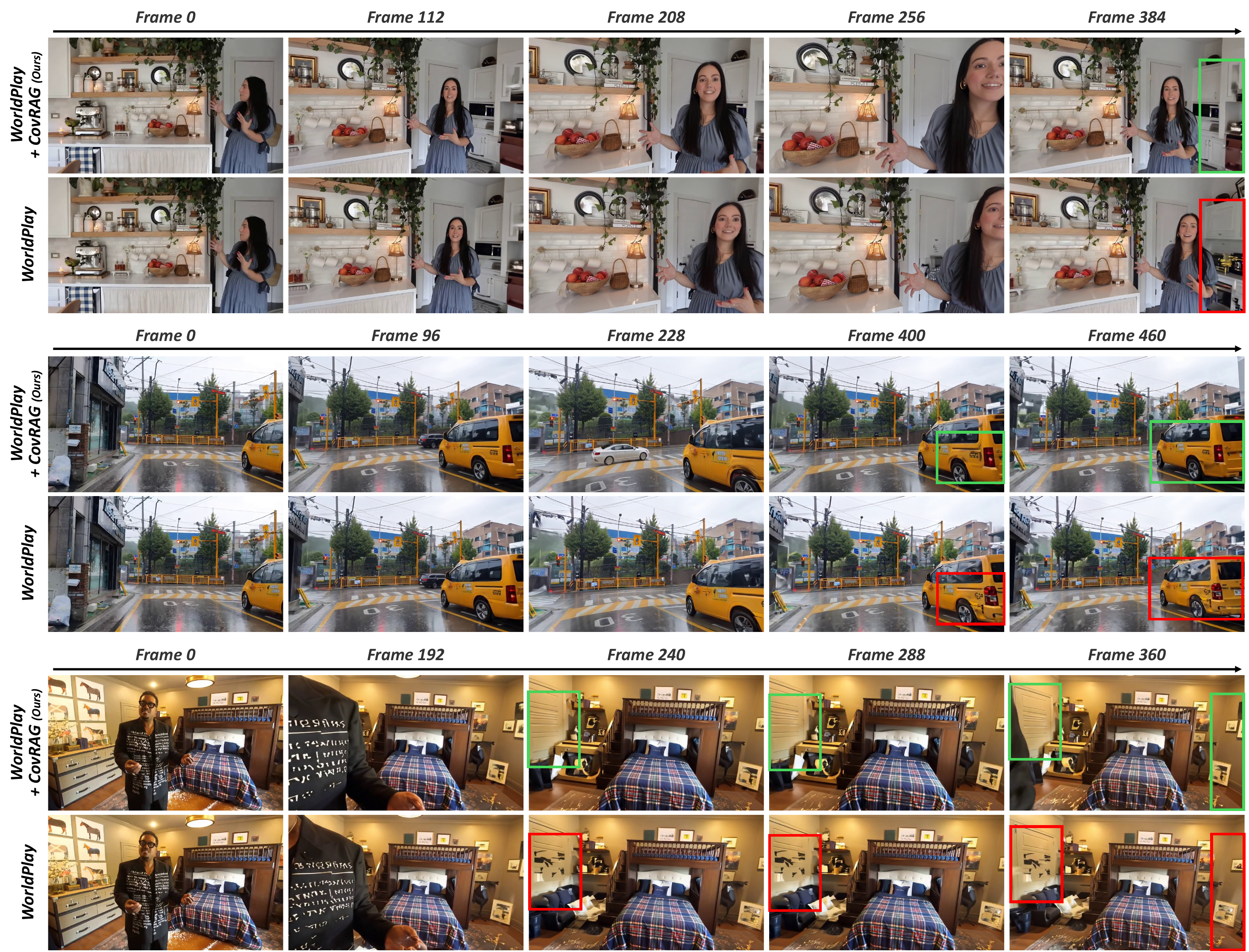}
    \caption{\textbf{Qualitative comparison in scenes containing motion.} Three examples from the dynamic-scene subset of SpatialVID. In each pair, the upper row uses WorldPlay-8B with \modelname, and the lower row uses its original FoV-based memory selection.
    The pretrained weights, generation settings, temporal context, and retrieval budget remain fixed.
    \textbf{\textcolor{ForestGreen}{Green}} and \textbf{\textcolor{red}{red}} boxes mark corresponding regions for comparing scene structure and appearance.}
    \label{fig:appendix_dynamic_qual}
\end{figure*}
\begin{table}[t]
  \centering
  \caption{\textbf{VBench evaluation in scenes containing motion.}
  We evaluate the same 100 generated SpatialVID rollouts as in Sec.~\ref{sec:dynamic_scenes}.
  Comparison between WorldPlay-8B's original FoV-based memory selection and \modelname under otherwise identical generation settings.}
  \label{tab:worldplay_vbench}
  \setlength{\tabcolsep}{3pt}
  \begin{tabular}{@{}lrrrrrr@{}}
    \toprule
    Method
      & \shortstack{Subject\\consistency$\uparrow$}
      & \shortstack{Background\\consistency$\uparrow$}
      & \shortstack{Motion\\smoothness$\uparrow$}
      & \shortstack{Dynamic\\degree$\uparrow$}
      & \shortstack{Aesthetic\\quality$\uparrow$}
      & \shortstack{Imaging\\quality$\uparrow$} \\
    \midrule
    WorldPlay-8B (FoV)
      & \textbf{0.9070} & 0.9019 & \textbf{0.99490}
      & 0.350 & \textbf{0.5197} & \textbf{0.7145} \\
    + \textbf{\modelname}
      & 0.9003 & \textbf{0.9026} & 0.99462
      & \textbf{0.440} & 0.5124 & 0.7075 \\
    \bottomrule
  \end{tabular}
\end{table}

\paragraph{Dynamic scenes.}
Figure~\ref{fig:appendix_dynamic_qual} compares \modelname with the original FoV-based memory selection of WorldPlay-8B~\cite{sun2026worldplay} on three SpatialVID~\cite{wang2026spatialvid} examples containing people or vehicles.
Both variants use the same pretrained model weights, generation settings, temporal context, and retrieval budget.
The highlighted regions illustrate better preservation of selected scene structures and appearance with \modelname across the shown views.
These examples demonstrate the usefulness of coverage-based retrieval in the evaluated scenes containing motion, although \modelname does not explicitly model object dynamics. 

Table~\ref{tab:worldplay_vbench} reports complementary VBench scores~\cite{huang2023vbench} on the same 100 generated videos evaluated in Sec.~\ref{sec:dynamic_scenes}.
\modelname increases dynamic degree from 0.350 to 0.440, while background consistency remains similar (0.9019 vs. 0.9026).
Subject consistency, motion smoothness, aesthetic quality, and imaging quality decrease slightly.
These mixed results complement the improvements in pose-revisit metrics and FVD reported in the main paper, rather than indicating a uniform improvement across all VBench metrics.

\begin{table}
\centering
\setlength{\tabcolsep}{5.5pt}
\renewcommand{\arraystretch}{1.08}
\caption{\textbf{Longer autoregressive rollouts on RealEstate10K.}
We evaluate generation horizons from $60$ to $120$ frames.
Lower is better for MEt3R, LPIPS, and FVD, while higher is better for PSNR.}
\resizebox{0.7\columnwidth}{!}{%
\begin{tabular}{c|l|cccc}
\toprule
\textbf{Frames} & \textbf{Method} & \textbf{PSNR $\uparrow$} & \textbf{MEt3R $\downarrow$} & \textbf{LPIPS $\downarrow$} & \textbf{FVD $\downarrow$} \\
\midrule
\multirow{2}{*}{$60$}
& WorldMem & 17.24 & 0.162 & 0.233 & 370.29 \\
& \textbf{\modelname} & \textbf{18.46} & \textbf{0.102} & \textbf{0.171} & \textbf{352.64} \\
\midrule
\multirow{2}{*}{$80$}
& WorldMem & 17.60 & 0.186 & 0.249 & 455.27 \\
& \textbf{\modelname} & \textbf{19.33} & \textbf{0.121} & \textbf{0.171} & \textbf{444.81} \\
\midrule
\multirow{2}{*}{$100$}
& WorldMem & 16.57 & 0.216 & 0.288 & 449.64 \\
& \textbf{\modelname} & \textbf{19.12} & \textbf{0.137} & \textbf{0.185} & \textbf{424.42} \\
\midrule
\multirow{2}{*}{$120$}
& WorldMem & 15.72 & 0.240 & 0.353 & 573.92 \\
& \textbf{\modelname} & \textbf{18.22} & \textbf{0.169} & \textbf{0.230} & \textbf{494.36} \\
\bottomrule
\end{tabular}%
}
\label{tab:supp_longer_rollout}
\end{table}
\begin{table}
\centering
\setlength{\tabcolsep}{5.5pt}
\renewcommand{\arraystretch}{1.08}
\caption{\textbf{Inference time and GPU memory usage.}
We report \modelname's retrieval latency per generation step, total inference time, peak GPU memory for rollout lengths from 40 to 120 frames.
Retrieval latency and memory remain nearly stable as the rollout length increases.}
\resizebox{0.8\columnwidth}{!}{%
\begin{tabular}{c|cccc}
\toprule
\textbf{Length} &
\textbf{\makecell{Retrieval latency\\per step (s)}} &
\textbf{\makecell{Total inference\\time (s)}} &
\textbf{\makecell{Peak memory\\(GB)}} &
\textbf{\makecell{Incremental time\\per frame (s)}} \\
\midrule
$40$  & 0.984 & 731.48  & 12.84 & -- \\
$80$  & 1.122 & 1579.05 & 12.89 & 21.21 \\
$120$ & 1.191 & 2426.90 & 12.98 & 21.16 \\
\bottomrule
\end{tabular}%
}
\label{tab:supp_inference_time}
\end{table}

\subsection{Quantitative Results}

\paragraph{Scalability as video length increases.}
\label{sec:vggt_dependency}
Specifically, starting from the first frame, we autoregressively generate 
$T\in\{60,80,100,120\}$ frames one frame at a time, and compare \modelname with WorldMem.
Table~\ref{tab:supp_longer_rollout} shows that \modelname consistently outperforms WorldMem across all tested horizons.
Both methods generally degrade as the number of generated frames increases, reflecting error accumulation in long autoregressive rollout.
However, \modelname remains better at every horizon.
From $60$ to $120$ frames, FVD increases by $55.0\%$ for WorldMem but by $40.2\%$ for \modelname, indicating smaller degradation over longer rollouts.

This experiment also provides end-to-end evidence regarding dependence on estimated geometry.
VGGT operates on autoregressively generated frames, so generation artifacts can affect subsequent geometry estimation, memory selection, and generation.
WorldMem provides a useful reference because its FoV-based selection rule does not depend on VGGT estimates, while using the same generator checkpoint and memory-conditioning architecture.
\modelname retains its advantage throughout the evaluated rollout lengths.

\paragraph{Inference efficiency.}
Table~\ref{tab:supp_inference_time} reports retrieval latency, total inference time, and peak GPU memory for rollout lengths from 40 to 120 frames.
Over this range, retrieval latency/step rises only from 0.98s to 1.19s, while peak memory stays nearly flat.
Although our retrieval step is slower than WorldMem, the increase in total video inference time is modest, averaging only 7.3\% from 40 to 120 frames, while yielding substantially better consistency, with 29.2\% lower MEt3R and 21.8\% lower LPIPS than WorldMem.

The sliding-window design bounds the number of frames processed in each VGGT inference call, while cached depth maps avoid repeatedly estimating geometry for the full history.
Coverage computation and frame selection can still depend on the number of historical candidates, so these empirical measurements do not imply constant cost for the entire retrieval pipeline.

\begin{table}[!t]
\centering
\setlength{\tabcolsep}{6pt}
\renewcommand{\arraystretch}{1.08}
\caption{\textbf{Sensitivity to controlled geometry perturbations.}
We evaluate \modelname under controlled perturbations on a subset of 100 RealEstate10K test videos.
\textit{Clean} denotes the original \modelname without perturbation.
Lower is better for MEt3R and LPIPS, while higher is better for PSNR.}
\resizebox{0.6\columnwidth}{!}{%
\begin{tabular}{l|c|ccc}
\toprule
\textbf{Setting} & \textbf{Lvl.} & \textbf{PSNR $\uparrow$} & \textbf{MEt3R $\downarrow$} & \textbf{LPIPS $\downarrow$} \\
\midrule
WorldMem & -- & 18.0 & 0.141 & 0.198 \\
\midrule
\textbf{\modelname} (\textit{Clean}) & -- & \textbf{18.6} & \textbf{0.100} & \textbf{0.158} \\
\midrule
\multirow{4}{*}{Image degradation}
& 1 & 18.3 & 0.115 & 0.175 \\
& 2 & 18.2 & 0.119 & 0.163 \\
& 3 & 18.2 & 0.122 & 0.180 \\
& 4 & 16.7 & 0.154 & 0.251 \\
\midrule
\multirow{2}{*}{Depth noise}
& 1 & 18.3 & 0.102 & 0.174 \\
& 2 & 18.2 & 0.119 & 0.176 \\
\bottomrule
\end{tabular}%
}
\label{tab:supp_geometry_robustness}
\end{table}
\paragraph{Robustness to inaccurate geometry.}
\modelname relies on estimated geometry to compute target-view coverage maps, so inaccurate depth estimates can affect memory selection.
However, geometry is used as a retrieval signal rather than as a rendered conditioning input to the generator.
VGGT-derived geometry is used to select historical RGB frames.
Estimated depth maps and geometric warps are not supplied as rendered conditioning inputs to the denoiser, and retrieved memories supplement the recent temporal context.

Table~\ref{tab:supp_geometry_robustness} evaluates the robustness of \modelname under inaccurate geometry.
Since \modelname uses estimated geometry as a retrieval signal for target-view coverage, errors in image-based geometry estimation or depth maps can affect memory selection.
Nevertheless, \modelname remains relatively stable under moderate perturbations and outperforms WorldMem in most settings.
Performance degrades noticeably only under stronger image degradation, where the geometry estimator receives substantially corrupted visual input.
Image degradations are applied only to the RGB inputs of VGGT; the RGB frames supplied to the generator remain unchanged.
For the perturbations reported in the table, image degradation levels L1--L4 use blur kernels $\{5,11,21,31\}$ and Gaussian noise standard deviations $\{0.01,0.03,0.05,0.08\}$.
Depth noise levels L1--L2 use scale biases $\{0.05,0.10\}$ and additive noise ratios $\{0.01,0.03\}$.
Together with the longer-rollout evaluation in Table~\ref{tab:supp_longer_rollout}, these results support tolerance to the tested moderate perturbations and effective retrieval over the evaluated autoregressive rollouts.
Catastrophic or temporally correlated geometry-estimation failures are not isolated by these experiments and remain a limitation.

\begin{figure}[t]
    \centering
    \includegraphics[width=0.9\columnwidth]{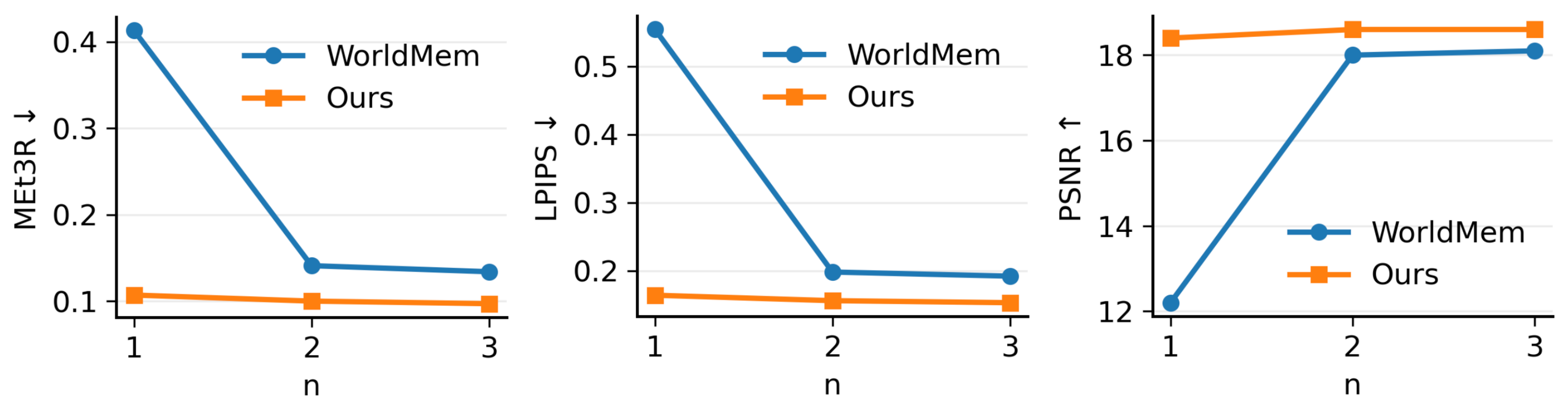}
    \caption{\textbf{Effect of retrieval budget on RealEstate10K.} We vary the number of retrieved memory frames $n$ and compare \modelname with WorldMem. \modelname consistently achieves better consistency across budgets, with marginal gains from $n=2$ to $n=3$.}
    \label{fig:budget_graph}
\vspace{-4mm}
\end{figure}
\begin{table}[t]
  \centering
  \caption{\textbf{Greedy vs. exhaustive memory selection.}
  With a retrieval budget of $n=2$, exhaustive search evaluates all
  pairs of $N$ candidate frames and maximizes residual coverage.
  Coverage/Opt.\ reports the mean per-retrieval ratio to the corresponding exhaustive optimum.
  Complexity is measured in $N$ at fixed coverage-map resolution,
  and time refers to frame selection.}
  \label{tab:greedy_vs_exhaustive}
  \begin{tabular}{@{}lcrrrr@{}}
    \toprule
    Selection
      & Complexity
      & Coverage/Opt.
      & \shortstack{Selection\\time (ms)$\downarrow$}
      & MEt3R$\downarrow$
      & PSNR$\uparrow$ \\
    \midrule
    Greedy (\modelname)
      & $\mathcal{O}(N)$
      & 94.2\%
      & \textbf{0.53}
      & \textbf{0.107}
      & 18.5 \\
    Exhaustive pair search
      & $\mathcal{O}(N^2)$
      & \textbf{100\%}
      & 2.33
      & 0.109
      & \textbf{18.6} \\
    \bottomrule
  \end{tabular}
\end{table}

\paragraph{Effect of retrieval budget.}
Figure~\ref{fig:budget_graph} shows the effect of varying the number of retrieved memory frames $n$.
\modelname consistently outperforms WorldMem across the plotted budgets and metrics, and achieves strong performance even with $n=1$.
This indicates that target-view coverage can identify informative memory frames even under a tight retrieval budget.
Increasing $n$ from 1 to 2 improves consistency, while the gains for \modelname from $n=2$ to $n=3$ are marginal.
These results support the use of a small retrieval budget and motivate our default choice of $n=2$.

\paragraph{Greedy vs. exhaustive memory selection.}
\label{sec:greedy_analysis}
We compare \modelname's greedy selection with exhaustive search to
assess its approximation quality and the effect on downstream generation.
With a retrieval budget of $n=2$, exhaustive search evaluates all
$\binom{N}{2}$ pairs of $N=|\mathcal{M}_t|$ candidate frames and finds the exact optimum
of the residual-coverage objective.
We evaluate on 100 RealEstate10K videos and measure frame-selection
time on an A6000 GPU.
Table~\ref{tab:greedy_vs_exhaustive} reports the mean of the ratios of
greedy residual coverage to the corresponding exhaustive optimum.
Greedy selection achieves a mean ratio of 94.2\%, while taking
0.53\,ms compared with 2.33\,ms for exhaustive search, a
$4.4\times$ speedup in selection.
For $n=2$ and a fixed coverage-map resolution, greedy selection scales
as $\mathcal{O}(N)$, whereas exhaustive pair search scales as
$\mathcal{O}(N^2)$.
The downstream results are similar: exhaustive search yields slightly
higher PSNR (18.6 vs. 18.5\,dB), while greedy selection yields slightly
lower MEt3R (0.107 vs. 0.109).
These results support greedy selection as an efficient approximation
with similar generation performance in this setting.

For a fixed target pose, initial temporal context, candidate pool, and coverage maps, our objective measures the union of target pixels covered beyond the temporal context.
Selecting no frames gives zero additional coverage.
Adding a frame cannot reduce coverage, and its additional gain cannot increase as the selected set grows.
The objective is therefore normalized, monotone, and submodular.
Under a retrieval budget of $n$ frames, greedy selection achieves at least $1-(1-1/n)^n$ of the optimal residual coverage~\cite{nemhauser1978analysis}.
This finite-budget bound is at least $1-1/e$ and equals $1-(1-1/2)^2=3/4$, or $75\%$, for our setting of $n=2$.
This is a worst-case guarantee for the estimated coverage objective within a single retrieval call, not for generated-video quality.

\section{Limitations and future work}
\paragraph{Limitations.}
\modelname assumes access to target camera poses and depends on a pretrained geometry estimator to compute target-view coverage.
Although the method retains an advantage under the tested moderate perturbations, inaccurate geometry can lead to misleading coverage estimates and suboptimal memory selection.
The longer-rollout experiments evaluate the complete autoregressive pipeline but do not isolate catastrophic or temporally correlated geometry-estimation failures.
Moreover, locally estimated depth maps do not guarantee globally consistent scene geometry.
\modelname also does not explicitly model object dynamics: large or rapid motion can make a geometrically relevant historical frame contain stale object states.
Severe generator artifacts or drift can likewise reduce the reliability of both estimated geometry and retrieved memories.

\paragraph{Future work.}
Our integrations with pretrained WorldPlay-8B and GEN3C demonstrate that coverage-based selection can be applied beyond the DFoT setting without retraining these generators.
More broadly, these results highlight the potential of coverage-based retrieval to strengthen long-term geometric consistency in video world models that support historical-frame conditioning, helping preserve coherent scenes over extended rollouts.
Future work can investigate motion-aware and uncertainty-aware coverage to account for stale observations and unreliable geometry.
Broader evaluations across scene types, camera trajectories, and generation architectures would further clarify the conditions under which coverage-based retrieval is most effective.

\section{Broader impacts}
This work aims to improve long-term consistency in video generation models. 
The proposed retrieval framework can benefit applications such as 3D content creation, embodied simulation, and interactive world modeling, where maintaining a coherent scene over time is important.
While the proposed method is primarily intended for advancing generative modeling, similar techniques could be misused for creating misleading visual content. We emphasize that responsible deployment and appropriate safeguards are necessary to mitigate such risks.

\clearpage

\end{document}